\documentclass{article}

\PassOptionsToPackage{numbers, compress}{natbib}

 \usepackage[main, final]{neurips_2026}

\usepackage[utf8]{inputenc} 
\usepackage[T1]{fontenc}    
\usepackage{hyperref}       
\usepackage{url}            
\usepackage{booktabs}       
\usepackage{amsfonts}       
\usepackage{nicefrac}       
\usepackage{microtype}      
\usepackage{xcolor}         
\usepackage{graphicx}
\usepackage{subcaption}
\usepackage{amsmath}
\usepackage{multirow}
\usepackage{multicol}
\usepackage{bbding}
\usepackage{wrapfig}
\usepackage{caption}
\usepackage{amssymb}
\usepackage{footnote}
\usepackage{enumitem}
\usepackage{algorithmic}
\usepackage{listings}
\usepackage{tabularx}
\usepackage {pifont} 
\usepackage{bm}
\usepackage{tablefootnote}
\usepackage{tikz}
\usepackage{makecell}
\usepackage{algorithm}
\usepackage{algorithmic}

\usepackage{mathtools}
\usepackage{amsthm}

\usepackage{cutwin}
\usepackage{colortbl}
\definecolor{cvprblue}{rgb}{0.21,0.49,0.74}

\definecolor{lightgray}{gray}{0.9}
\definecolor{linecolor}{rgb}{0.82, 0.94, 0.75}
\definecolor{mamba}{RGB}{153, 151, 239}
\definecolor{kaiming-green}{RGB}{57,181,74} 
\definecolor{pretty-blue}{RGB}{0, 113, 188}

\definecolor{up}{RGB}{68,169,32}
\definecolor{down}{RGB}{255,0,0}
\definecolor{linecolor}{gray}{.65}

\usepackage{authblk}

\AtEndPreamble{
    \usepackage[capitalize]{cleveref}
    \crefname{section}{Sec.}{Secs.}
    \Crefname{section}{Section}{Sections}
    \crefname{table}{Tab.}{Tabs.}
    \Crefname{table}{Table}{Tables}
}

\hypersetup{
    colorlinks=true,
    linkcolor=red,
    filecolor=magenta,      
    urlcolor=magenta,
    citecolor=kaiming-green
}

\theoremstyle{plain}

\theoremstyle{definition}

\theoremstyle{remark}

\usepackage[textsize=tiny]{todonotes}

\title{PixelPonder: Dynamic Patch Adaptation for Enhanced Multi-Conditional Text-to-Image Generation}

\author{%
  \textbf{Yanjie Pan}$^{1*}$
  ~~ \textbf{Qingdong He}$^2$\thanks{Equal contributions.}
  ~~ \textbf{Zhengkai Jiang}$^2$
  ~~ \textbf{Pengcheng Xu}$^3$
  ~~ \textbf{Chaoyi Wang}$^4$ 
  ~~ \textbf{Jinlong Peng}$^2$
  ~~ \textbf{Haoxuan Wang}$^1$
  ~~ \textbf{Yun Cao}$^2$
  ~~ \textbf{Zhenye Gan}$^2$
  ~~ \textbf{Mingmin Chi}$^1$
   ~~ \textbf{Bo Peng}$^{1\dagger}$
  ~~ \textbf{Yabiao Wang}$^{5,2}$\thanks{Corresponding author.}
   \\
  \normalsize $^1$Fudan University ~~ $^2$Tencent Youtu Lab ~~ $^3$Western University \\
  ~~ $^4$University of Chinese Academy of Sciences ~~ $^5$Zhejiang University
} 

\begin{document}

\maketitle

\begin{figure}[h]
    \centering
    \includegraphics[width=1\linewidth]{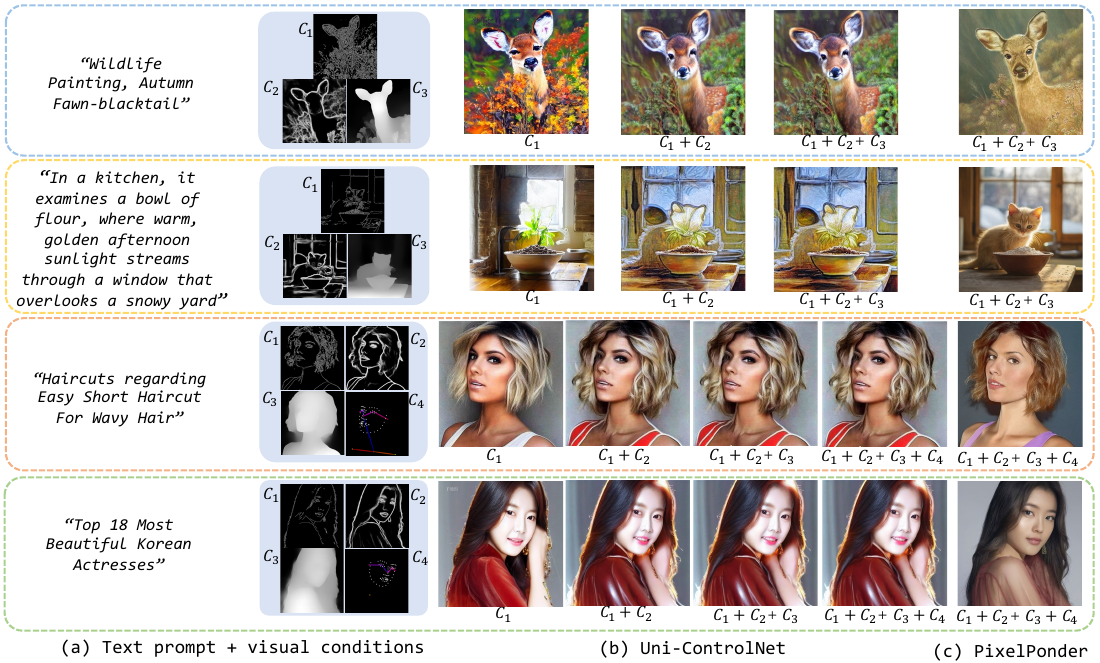}
    \caption{\textbf{Visualization comparison between Uni-ControlNet~\citep{zhao2024uni} and our proposed method in different conditional controls with the same text prompt.} 
        \textit{\textbf{(a, left two columns)}} Text and various visual controls, where $C_{1}$, $C_{2}$, $C_{3}$ and $C_{4}$ denotes canny, sketch, depth and pose map respectively. 
        \textit{\textbf{(b, middle three or four columns)}} Generation results from UniControlNet. 
        \textit{\textbf{(c, last column)}} Generation results from our PixelPonder. 
        Previous methods struggled to generate coherent results under multiple conditions, while our results maintain strong similarity to the respective visual controls.}
    \label{fig:compare}
\end{figure}

\begin{abstract}
Recent advances in diffusion-based text-to-image generation have demonstrated promising results through visual condition control. However, existing ControlNet-like methods struggle with compositional visual conditioning - simultaneously preserving semantic fidelity across multiple heterogeneous control signals while maintaining high visual quality, where they employ separate control branches that often introduce redundant guidance during the denoising process, leading to structural distortions and artifacts in generated images. To address this issue, we present PixelPonder, a novel unified control framework, which allows for effective control of multiple visual conditions under a single control structure. Specifically, we design a patch-level adaptive condition selection mechanism that dynamically prioritizes spatially relevant control signals at the sub-region level, enabling precise local guidance without global interference. Additionally, a time-aware control injection scheme is deployed to modulate condition influence according to denoising timesteps, progressively transitioning from structural preservation to texture refinement and utilizing the control information from different categories to promote finer image generation. Extensive experiments demonstrate that PixelPonder surpasses previous methods across different benchmark datasets, showing superior improvement in spatial alignment accuracy while maintaining high textual semantic consistency. The code and models are available at \url{https://hithqd.github.io/projects/PixelPonder/}.
\end{abstract}
   
\section{Introduction}
The emergence of diffusion-based text-to-image synthesis~\citep{rombach2022high,dhariwal2021diffusion,peebles2023scalable,podell2023sdxl,ramesh2021zero,ramesh2022hierarchical} have driven significant progress in single-modality conditioned generation frameworks~\citep{ye2023ip,li2025controlnet,mou2024t2i,cao2024controllable,li2023gligen}. While these approaches demonstrate enhanced controllability through visual-textual alignment, fundamental limitations persist due to modality-specific biases in visual conditioning. As evidenced in~\citep{si2024freeu,he2024dynamiccontrol}, distinct visual control signals exhibit complementary characteristics, such as depth maps effectively regulate inter-object spatial relationships but lack fine-grained object details, whereas canny maps capture precise texture contours while disregarding global structural context. This modality specialization gap raises a critical challenge: \textit{\textbf{How to establish an effective fusion paradigm for multi-visual condition control that synergistically integrates heterogeneous visual features while resolving inter-modality conflicts?}}

\begin{wrapfigure}{r}{0.49\linewidth}
    \centering
    \includegraphics[width=1\linewidth]{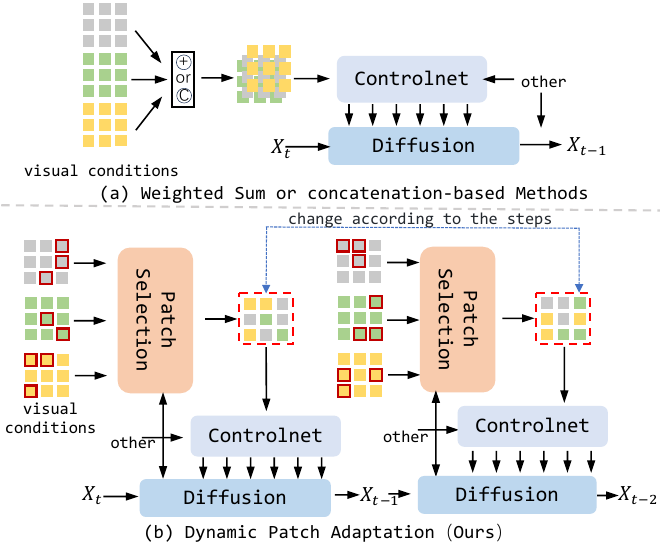}
    \vspace{-1.5em} 
    \caption{\textbf{Comparison of different schemes dealing with multiple condition.} \textbf{(a)} Existing methods integrate various visual controls into a highly consistent visual control signal in the temporal domain through concatenation or weighted summation. \textbf{(b)} We employ dynamic patch selection to reconstruct visual conditions at the patch level, enabling multi-visual controllable generation in the generative model.}
    \label{fig:method_compare}
\end{wrapfigure}
In practical applications, it is often necessary to describe the visual features of a significant object using multiple visual conditions to achieve precise control over its generation. Users typically desire to simultaneously control both the layout and the intricate details. However, visual conditions that are rich in layout and detail information are challenging to integrate into a single visual condition map. A common scenario involves using canny maps to represent details and depth maps to control the layout. In such case, these visual conditions are highly overlapping and complementary to one another.

Bridging the redundant information among these visual conditions and reasonably utilizing the emphasis of various visual conditions to coordinate the generation of the object is exactly the issue that this paper wants to focus on. As shown in Fig.\ref{fig:method_compare}(a), prior efforts including Cocktail~\citep{hu2023cocktail} and Uni-ControlNet~\citep{zhao2024uni} propose to consolidate various visual conditions into unified latent representations. As illustrated in Fig.\ref{fig:compare}, generating normal images under multi-modal and coupled visual condition controls remains a difficult task for these approaches. While this coarse-grained unification alleviates inter-condition conflicts through dimensionality reduction, it remains suboptimal in preserving modality-specific attributes critical for precise control (e.g., high-frequency textures in cannys vs. low-frequency structures in depth). The inherent information bottleneck induced by latent compression disproportionately attenuates complementary cross-modal features, fundamentally limiting their capacity to synergize discriminative visual cues. This constitutes a critical limitation: current unification paradigms prioritize conflict mitigation over inter-modal feature reinforcement.

To address the aforementioned challenges, we introduce an innovative visual condition combination method termed \textit{PixelPonder}. As illustrated in Fig.\ref{fig:method_compare}(b), our method distinguishes itself from previous approaches by integrating diverse visual conditions at the patch level. Recognizing that visual conditions enriched with low-frequency or high-frequency signals significantly influence various denoising stages during the image generation process~\citep{ho2020denoising}, we have developed an adaptive patch selection mechanism that dynamically adjusts with timesteps. This mechanism decouples each visual condition in the temporal domain, enabling the use of distinct combinations of visual conditions to regulate the denoising process at different stages. Our proposed combination method amalgamates various visual conditions with greater granularity, and facilitating complementary control of high- and low-frequency information within the temporal domain. 

It is worth noting that when operating under a single visual condition control, our proposed method effectively reverts to the single visual control scheme. The efficacy of ControlNet's single visual condition control has been extensively validated in numerous studies~\citep{zhang2023adding, zhao2024uni, qin2023unicontrol}. Therefore, this paper does not investigate the generative capabilities of our method under single visual condition scenarios. Our main contributions are summarized as follows:
\begin{itemize}
  \item \textbf{\emph{New Insight:}} We reveal that current multi-visual condition control methods perform poorly in terms of generation quality and controllability, making it difficult for generated images to align with various visual conditions and lacking effective solutions for improvement.
  
  \item \textbf{\emph{Efficient Patch Adaptation:}} We propose PixelPonder, a novel mechanism which can refine the combination of multi-visual conditions by transitioning from the image level to the patch level, thereby enabling finer-grained controllable generation.
  
  \item \textbf{\emph{Flexible Condition Combination:}} Our PixelPonder supports flexible visual control combination, addressing the redundant information conflict of multiple visual conditions for the same object.

  \item \textbf{\emph{Promising Results:}} We provide a consolidated and public evaluation of controllability and fidelity under various conditional controls, and demonstrate that PixelPonder comprehensively outperforms existing methods.

\end{itemize}

\section{Related Work}
\label{related_work}

\subsection{Diffusion-based Generative Models}
Diffusion models~\citep{sohl2015deep, ho2020denoising, nichol2021glide, podell2023sdxl,ramesh2021zero, chen2023pixart} have made remarkable strides in the field of image generation. Thanks to improvements in sampling strategies~\citep{ho2020denoising,song2020denoising,song2020score},  as well as the application of latent diffusion methods~\citep{rombach2022high}, the computational resources required for these models to generate high-quality images have been substantially reduced, demonstrating promising prospects in various applications~\citep{cao2024controllable, brooks2023instructpix2pix, hertz2022prompt, ramesh2022hierarchical}. In the domain of text-to-image generation, diffusion models~\citep{rombach2022high} combined with U-Net~\citep{ronneberger2015u} have successfully integrated text control into the image generation process through cross-attention mechanisms. To further enhance generative capabilities, the Transformer architecture has been integrated into diffusion models~\citep{peebles2023scalable,chen2024pixart,esser2024scaling}. Building on this foundation, FLUX~\citep{black2024flux}, which introduces flow matching objectives, has achieved state-of-the-art performance in image generation.

\subsection{Controllable Text-to-Image Diffusion Models}
The generation of images controlled by textual descriptions often lacks spatial control. To enhance spatial control capabilities, numerous studies~\citep{mou2024t2i, zhang2023adding, li2023gligen, huang2023composer, cheng2023layoutdiffuse, yang2023reco, zhang2023controllable} have focused on incorporating visual control signals into the generation process. In this context, ControlNet~\citep{zhang2023adding} proposed a spatial control network architecture, encoding additional visual conditions into latent representations and injecting control signals into the corresponding backbone network using zero convolution, thus achieving a spatial alignment mechanism within diffusion models. Due to its straightforward approach and effective control results, it has been widely adopted in various text-to-image diffusion models~\citep{rombach2022high,black2024flux}. Furthermore, recent research has used prompt engineering techniques~\citep{li2023gligen,yang2023reco} and inter-attention constraints~\citep{chen2024training,xie2023boxdiff} to facilitate more orderly generation.

Some studies~\citep{qin2023unicontrol, mou2024t2i, wang2024instancediffusion, bar2023multidiffusion} have addressed the differences and diversity of various visual conditions in different modalities using the mix of experts (MOE) to decouple different visual conditions into distinct feature subspaces, thus establishing a unified visual control architecture. Subsequently, other research has explored the possibility of generating multiple condition controls within a single diffusion model through simple feature aggregation methods~\citep{hu2023cocktail,zhao2024uni}. However, these methods are based on the low coupling of various conditions, which limits their effectiveness in tasks requiring the control of complementary conditions.
\vspace{-2mm}
\section{Methodology}\label{method}
\begin{figure*}[t]
\centering
\includegraphics[width=\textwidth]{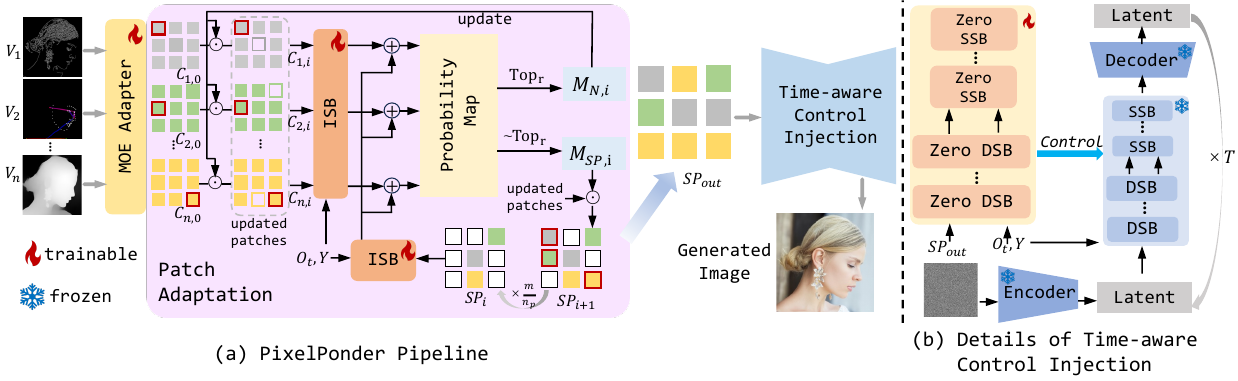}
\caption{\textbf{Overall pipeline of the proposed}. The Patch Adaptaiom Module segments various visual conditions \(\{V_1, V_2, \ldots, V_n\}\) into patches and dynamiclly reorganizes the visual conditions at the patch level to obtain a unified visual control signal \(SP_{out}\). Subsequently, \(SP_{out}\) is sent to the control network, with the Time-aware Control Injection scheme to achieve high-quality image generation under multiple visual conditions.
}
\label{fig:framework}
\end{figure*}

In this paper, we introduce PixelPonder, a specialized framework for handling complex scenarios of image generation under multiple visual conditions. The pipeline of PixelPonder is demonstrated in Fig. \ref{fig:framework}. In this section, we first provide a brief overview of the previous Diffusion Transformer in Section~\ref{subsec:preliminary}, along with ControlNet. Then, in Section~\ref{subsec:patch selection}, we dive into a detailed description of the Patch Adaptation Module, which primarily focuses on the autoregressive reorganization of visual conditions at the patch level. In Section ~\ref{subsec:time-aware}, we elaborate on the time-step awareness injection strategy and the control implementation of the control network. Finally, we summarize our training optimization objectives in Section~\ref{subsec:training}.
\subsection{Preliminary}
\label{subsec:preliminary}

\textbf{Diffusion Transformer} (DiT)\citep{vaswani2017attention, peebles2023scalable} is applied in models such as FLUX ~\cite{black2024flux} and Stable Diffusion 3~\citep{esser2024scaling}. Compared to the Latent Diffusion Model~\citep{rombach2022high}, these models utilize a transformer~\citep{ruan2023mm} instead of U-Net as the denoising network to iteratively process noise tokens.

A model with a DiT architecture handles noisy image tokens \( X \in \mathbb{R}^{N \times d} \) and text condition tokens \( Y \in \mathbb{R}^{M \times d} \), where \( d \) is the dimension of the embedding layer, \( N \) and \( M \) represent the number of tokens for images and text, respectively. Additionally, they redefine the mapping from images to noise using flow matching. In this rectified mapping, the data at time \( t \), denoted as \( z_t \), is transformed from the real image \( x_0 \in \mathbb{R}^{C \times H \times W} \) to pure noise \( N \sim \mathcal{N}(0, I) \) at the following position:
\begin{equation}
    z_t = (1-t)x_0+tN,
\end{equation}
\begin{equation}
    \vec{V} : \mathcal{N}\ \rightarrow \mathcal{Z}\ ,
\end{equation}
where \(\mathcal{N}\) represents the spatial distribution of noisy images. \(\mathcal{Z}\) represents the spatial distribution of real images. \(\vec{V}\) is the vector field from \(\mathcal{N}\) to \(\mathcal{Z}\). 

In FLUX, a series of flow matching modules are included, among which the denoising backbone network consists of the Double Stream Block (DSB) and the Single Stream Block (SSB).

\noindent \textbf{ControlNet}~\citep{zhang2023adding} is designed to inject visual control information into a denoising network, thereby guiding the iterative process of noise tokens. Specifically, ControlNet freezes the parameters \( \Theta\) of the backbone network \(F(\cdot;  \Theta)\) and creates a trainable copy \( \Theta_c\) of these parameters to incorporate zero-initialized convolutional layers \(Z\). When applied to generative models based on the U-Net architecture, a common practice is to add the visual condition \(C_i\) to the noise \(x_t\) as input to the trainable copy, while utilizing other signals \(s\) that is consistent with the backbone network. This process can be formalized as follows:
\begin{equation}
    y = F(x_t, s, \Theta) + Z(F(x_t +M(C_1, \ldots, C_n), s,\Theta_c)),
\end{equation}
where \(M\) serves as a unified representation of various visual condition fusion methods, $n$ represents the category id.

\subsection{Patch Adaptation}
\label{subsec:patch selection}
\begin{wrapfigure}{r}{0.5\linewidth}
\vspace{-8mm}
\begin{minipage}{0.95\linewidth}
\begin{algorithm}[H]
        \caption{Patch Adaptation}
        \label{alg:1}
        \textbf{Input:} visual condition features \(C_n \in \mathbb{R}^{N \times d}\), time step information \( O_t \) and textual feature \( Y \).\\
        \textbf{Output:} the unified visual control $SP_{\text{out}}$ 
    
    \begin{algorithmic}[1] 
    \STATE \( SP_0 = \textbf{0}_{N \times d} \)
    \STATE $ C_n = [P_n^0, \ldots, P_n^m]$ 
    \STATE $ \mathcal{C}_0 = \{ C_1, C_2, \ldots, C_n \}$
    \STATE $i \leftarrow 0$
    \STATE $n_p \leftarrow $ number of selected patches
    \WHILE{$i \cdot n_p < m$} 
    \STATE $M_{\text{prob}} = [\text{ISB}_{\text{C}}(\mathcal{C}_i, O_t, Y)] + \text{ISB}_{\text{SP}}(SP_i, O_t, Y)$
    \STATE $M_{N,i} = \text{Zero}(\text{Top}_r(M_{prob}))$, $M_{sp,i} = \sim M_{N,i}$
    \STATE $ \mathcal{C}_{i+1} = \mathcal{C}_{i} M_{N,i} $
    \STATE $ SP_{i+1} = SP_{i} + M_{sp,i} \mathcal{C}_{i} $
    \STATE $i \leftarrow i+1$
    \ENDWHILE
    \STATE \textbf{return} $SP_i$
\end{algorithmic}
\end{algorithm}
\end{minipage}
\end{wrapfigure}
The Patch Adaptation Module (PAM) aims to recompose various visual conditions at the patch level into a unified visual condition. This is achieved through an autoregressive iterative combination process that combines patches between different visual conditions. The specific algorithm flow of PAM is shown in algorithm \ref{alg:1}.

Specifically, at time step \( t \), the PAM initializes a zero vector unified control condition \( SP_0 \in \mathbb{R}^{N \times d} \) to obtain the final unified visual control combination output \( SP_{\text{out}} \). The combination selection is defined as an update process \( U_i \) from \( SP_i \) to \( SP_{i+1} \), with $i$ representing the index of the number of iterations in the update process. This iterative update process can be formalized as follows:
\begin{equation}
    SP_{i+1} = U(SP_{i}, \ldots, O_t),
    \label{eq4}
\end{equation}
where \(O_t\) refers to the time step information, and its specific role will be discussed in next subsection.

Assuming that the visual input to the PAM consists of \( N \) types of condition images, denoted as \( V_1, V_2, \ldots, V_n \), we use the corresponding visual condition feature encoders \( E_i \) to obtain various visual condition features \( \mathcal{C} = \{ C_1, C_2, \ldots, C_n \} \), \(C_k \in \mathbb{R}^{e \times d}\) and \(d\) denotes the feature dimension. To analyze and process the visual condition features in a more fine-grained manner, we can consider the visual condition feature \( C_k \) as represented by patches \(P_i^m \in \mathbb{R}^{d}\), expressed as:
\begin{equation}
    C_k = [P_k^0, \ldots, P_k^m],
\end{equation}
where \( P \) refers to a patch, \( k \) denotes the random category id, and \( m \) denotes the number of patches.

In an update process \( U_i \), the PAM selects a set \(  \mathcal{P}_i = \{ P_{n,i}^s \mid s\in[0, m], k\in[0, n] \} \) composed of multiple patches \( P_{n,i}^s \) from various visual characteristics \(  \mathcal{C}_i \). For convenience, we define this selection process as \( F_{\text{sp}} \). The composition of this set \(  \mathcal{P}_i \) considers two aspects of information: \textit{\textbf{a) The dominant role of textual information \( Y \) in the current time step in image generation}} , and \textit{\textbf{b) The influence of previously selected patches \( \mathcal{P}_0, \ldots, \mathcal{P}_{i-1} \) on the overall composition of image control.}}  The update process from \( SP_i \) to \( SP_{i+1} \) can be more specifically expressed as:
\begin{equation}
    U_i(SP_{i}, \ldots, O_t) = SP_{i} + \mathcal{P}_i,
\end{equation}
\begin{equation}
\mathcal{P}_i = F_{\text{sp}}(\mathcal{C}_i, \mathcal{P}_0, \ldots, \mathcal{P}_{i-1}, O_t, Y).
\end{equation}

Specifically, to achieve \( F_{\text{sp}} \), we design the Image Stream Block (\textbf{ISB}). To ensure that the textual control signal remains consistent globally, 
we change the DSB text input selection strategy, abandoning the textual information \( Y_{\text{DSB}} \) generated in dual-stream matching as input and instead using a globally unified textual feature representation \( Y \) as the input textual information.
We modified the DSB text input selection strategy to adopt a globally unified textual feature representation \( Y \) as the input textual information.
The ISB is used to obtain the weights of each patch in various visual condition features under the current textual control \( Y \) at time step \( t \), denoted as \(  \mathcal{W}_c = \{ W_{0,i}, \ldots, W_{n,i} \} \). Furthermore, to meet requirement \textit{\textbf{b)}} of \( F_{\text{sp}} \), we apply an ISB to \( SP_i \) outside of \(  \mathcal{C}_i \) to obtain a global weight \( W_{\text{sp}} \) that represents the global composition information of \( SP_i \) under the current \( U_i \). During training, the Gumbel-Softmax ~\citep{jang2016categorical} is adopted to ensure the differentiability of the Top‑r selection operation.
The final obtained patch selection probability map \( M_{\text{prob}} \) is given by:
\begin{equation}
    M_{\text{prob}} = [ \mathcal{W}_c] + W_{\text{sp}},
\end{equation}
\begin{equation}
\mathcal{W}_c = \text{ISB}_{\text{C}}(\mathcal{C}_i, O_t, Y), \quad W_{\text{sp}} = \text{ISB}_{\text{SP}}(SP_i,O_t, Y).
\end{equation}

\begin{wrapfigure}{r}{0.69\linewidth}
    \centering
    \includegraphics[width=1\linewidth]{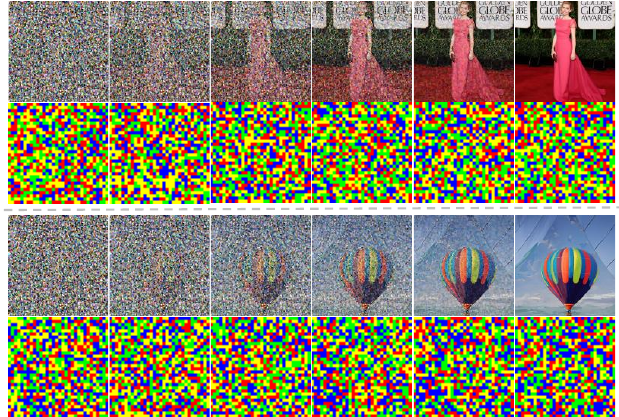}
    \caption{\textbf{Denoising results at steps 25, 20, 15, 10, 5, and 1}, where each pair includes the generated result and the selection results \(SP_{out}\) at the current time step. In each \(SP_{out}\), canny, pose, depth and sketch are indicated in red, blue, green, and yellow respectively.}
    \label{fig:patch_selection}
\end{wrapfigure}
In each iteration, to mask the influence of previously selected patch blocks \(\{P_0, \ldots, P_{i-1}\}\) on the combination selection process \(U_i\), we use a binary mask \(M_{k,i}\) to update \(C_{k}^{i-1}\) to obtain \(C_{k}^{i}\). The binary mask \(M_{k,i}\) is derived from \(M_{prob}\):
\begin{equation}
M_{N,i} = \text{Zero}(\text{Top}_r(M_{prob})),
\end{equation}
\begin{equation}
[\bigcup_{k=1}^{N} M_{k,i}] = M_{N,i},
\end{equation}
where \(\text{Top}_r\) retrieves the positions of the top \(r\) patches with the highest probabilities in \(M_{prob}\), and \(\text{Zero}\) sets the corresponding positions in the binary mask to zero.


Fig.\ref{fig:patch_selection} illustrates the visualization results of \(SP_{out}\) obtained by PAM in different time steps, as well as the generated images. In the denoising process, the generated images exhibit a noticeable tendency towards patchiness. The segmentation contours between the generated image patches are prominent. 
Both examples demonstrate the characteristics of PAM: a) PAM exhibits a balanced tendency in its selections over various time steps. b) For patches at the same location, the global selections include patches from all categories.

It is noted that \(C_{n}^{0}\) is equivalent to \(C_{n}\), which means that we do not process \(C_{n}\) in \(U_{0}\). The representation of the visual condition \(C_{n}\) during the iteration process is as follows:
\begin{align}
    C_{k}^{i} = 
    \begin{cases} 
    \mathbf{E}_k(V_k) & \text{if } i = 0 \\ 
    M_{k,i}  C_{k}^{i-1} & \text{if } i > 0 .
    \end{cases}
\end{align}

While obtaining \(M_{N,i}\), we can also acquire the mask \(M_{sp,i}\) corresponding to the selected patch \(P_i\) during the update process \(U_i\):
\begin{equation}
M_{sp,i}=\sim M_{N,i}, \mathcal{P}_i = M_{sp,i} \mathcal{C}_i ,
\end{equation}
where \(\sim\) represents the negation operation.

\begin{figure*}[t]
\centering
\includegraphics[width=0.95\linewidth]{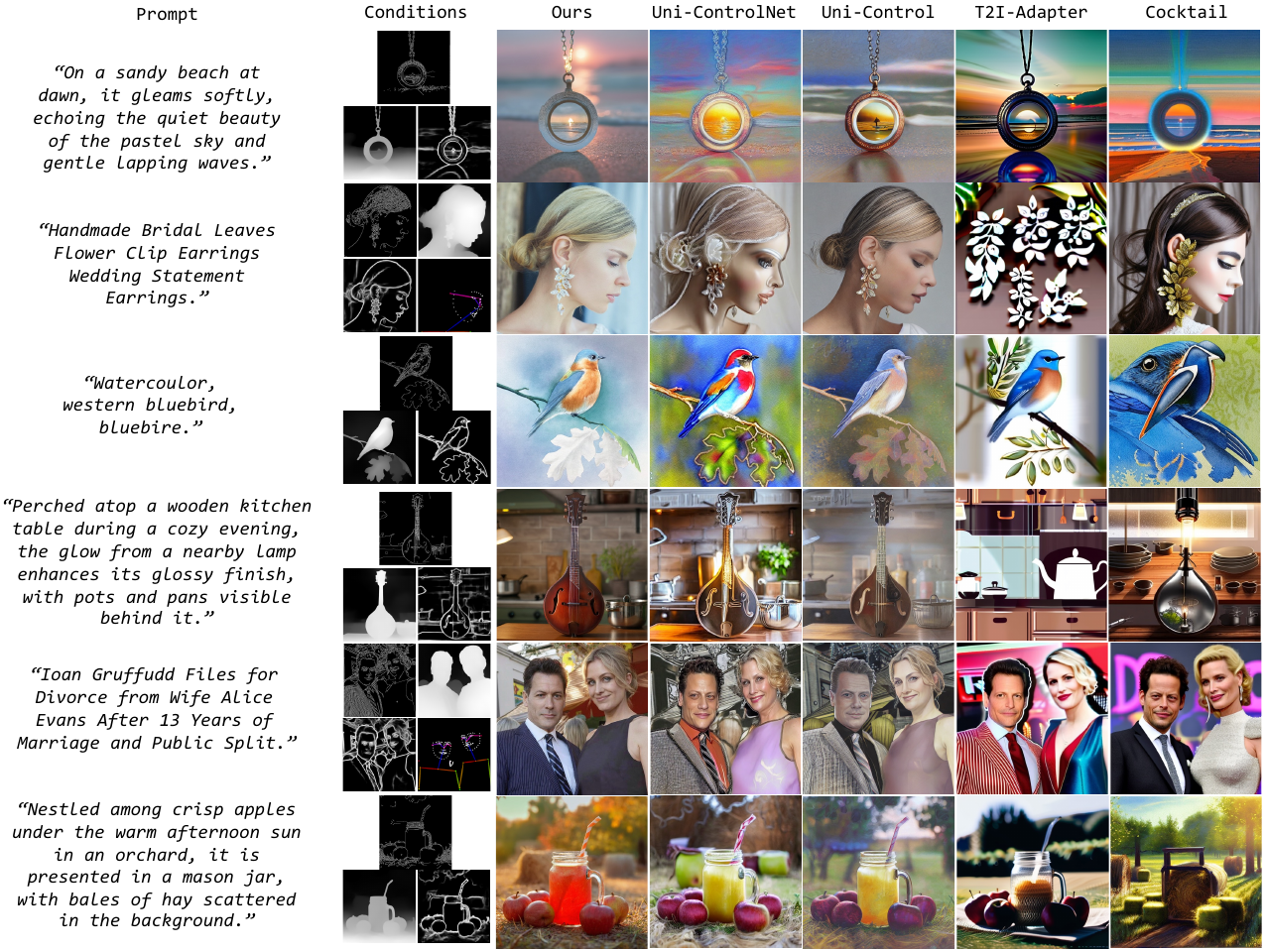}
\caption{\textbf{Visualization comparison} between official methods and our proposed model in different conditional controls.}
\label{fig:vis_compare}
\end{figure*}

\subsection{Time-aware Control Injection}
\label{subsec:time-aware}
The time-aware control injection is designed to fully utilize the visual control effects of various visual conditions at different time steps during the denoising process. This is achieved by introducing additional temporal information \(O_t\) in the PSM at each time step, which controls the patch selection tendency in each update process \(U\). To ensure that the patch selection tendencies in all selection processes \(U\) within a time step are consistently influenced by the temporal information, we use a unified temporal signal \(O_t\) at the current time step for control, as shown in Eq. \ref{eq4}.

Specifically, we introduce the time-step information \(O_t\) into the visual conditions and the global ISB module. This allows the ISB to simultaneously acquire time-step control information to obtain patch selection weights that are more aligned with the denoising temporal flow, which can be formalized as follows: 
\begin{equation}
Control = D'(SP_{out},Y,O_t,\Theta_c),
\end{equation}
\begin{equation}
I_r = D(N,Control,Y,\Theta) ,
\end{equation}
where \(D\) denotes the original model, \(\Theta\) indicates the model parameters, \(D'\) refers to the control network, and \(\Theta_c\) signifies the model parameters of the control network, \(I_r\) represents the generated image.
Considering the large number of parameters in the model, we reduce the number of modules in the control network and align the image streams sequentially to the corresponding backbone modules.

To achieve stable control of visual composition under time awareness, we initialize the last linear layer to zero from the ControlNet and FLUX, denoted as zero-initialized Double Stream Block (\textbf{Zero DSB}) and zero-initialized Single Stream Block (\textbf{Zero SSB}), which can be formulated as:
\begin{equation}
   Attention(Q,K,V) = Z( (\frac{QK^T}{\sqrt{d_{\mathrm{k}}}}),V ).
\end{equation}

\subsection{Training}
\label{subsec:training}
\noindent\textbf{Objective.}
Given a set of multi-visual prompts \( \mathcal{C} \) and a text prompt \( Y \), we fine-tune our method to reconstruct the original image \( x \). During the training process, we freeze the backbone network and only optimize the patch selection network and the zero controllable network. Our training objective combines \( V \) and \( V' \) to obtain the visual representation \( V_{n \rightarrow c} \) under the visual prompts \( C \). This encourages the visual control network to learn the internal visual condition control flow \( V' \) based on the \( V \) pre-learned by the backbone network. The loss function \(\mathcal{L}\) is formalized as: 
\begin{equation}
\mathcal{L} = \| (N-x_0) - \int_{0}^{1} \vec{V} \oplus \vec{V}'(z_t, \mathcal{C}, Y, t)dt \|^2 ,
\end{equation}
where the \( \oplus \) represents the vector sum of two fields.

\section{Experiments}
\label{experiments}
We validate the effectiveness of PixelPonder on the MultiGen-20M~\citep{li2025controlnet} and Subject-200K\citep{tan2024ominicontrol} datasets. Our evaluation primarily focuses on several leading methods, including T2I-Adapter~\citep{mou2024t2i}, Uni-ControlNet~\citep{zhao2024uni}, UniControl~\citep{qin2023unicontrol}, and Cocktail~\citep{hu2023cocktail} in
the realm of multiple controllable text-to-image diffusion models. It is worth noting that we conducted experiments on Cocktail using only pose and sketch maps, as it only supports pose, sketch, and segmentation maps. Although the models of other approaches such as AnyControl~\cite{sun2024anycontrol} are public, their code cannot be successfully run after many attempts. More implementation details can be found in the supplementary material.

\begin{table*}[t]
\caption{%
    Comparison on the MultiGen-20M~\citep{li2025controlnet} and Subject-200K~\citep{tan2024ominicontrol} datasets. The best and second-best results are highlighted in \textbf{bold} and \underline{underline}, respectively. In the \textbf{Conditions} column, ``all'' denotes the use of canny maps, sketches, pose maps, and depth maps simultaneously during generation.
}
\resizebox{\linewidth}{!}{
\begin{tabular}{c|c|c|c|c|c|c|c|c|c|c}
\toprule
\multicolumn{1}{c|}{} & \multicolumn{1}{c|}{}  & \multicolumn{4}{c|}{\textbf{MultiGen-20M}} & \multicolumn{4}{c|}{\textbf{Subject-200K}} & \multicolumn{1}{c}{}   \\ \cline{3-6} \cline{7-10} 
\multicolumn{1}{c|}{\multirow{-2}{*}{\textbf{Methods}}}& \multicolumn{1}{c|}{\multirow{-2}{*}{\textbf{Conditions}}} &  \multicolumn{1}{c|}{\textbf{FID ($\downarrow$)}} & \multicolumn{1}{c|}{\textbf{CLIP Score~($\uparrow$)}} & \multicolumn{1}{c|}{\textbf{SSIM~($\uparrow$)}} & \multicolumn{1}{c|}{\textbf{MUSIQ ($\uparrow$)}} &  \multicolumn{1}{c|}{\textbf{FID ($\downarrow$)}} & \multicolumn{1}{c|}{\textbf{CLIP Score~($\uparrow$)}} & \multicolumn{1}{c|}{\textbf{SSIM~($\uparrow$)}} & \multicolumn{1}{c|}{\textbf{MUSIQ ($\uparrow$)}} & \multicolumn{1}{c}{\multirow{-2}{*}{\textbf{Inference Time ($\downarrow$)}}}\\ 
\midrule
T2I-Adapter      &all     &66.95                  &71.47                  &24.03                  &57.95                  &64.72                  &74.35                  &31.76                  &55.07  & 9.33\\
Uni-ControlNet  &all     &32.58                  &78.08                  &29.37                  &65.85                  &44.35                  &\underline{77.40}&37.98                  &66.75  & 7.22\\
UniControl &all     &25.15                  &74.09                  &\underline{35.58}&\textbf{72.05} &\underline{30.95}&72.96                  &\underline{47.31}&\textbf{67.50}  & 7.39\\ 
Cocktail     &pose+hed&\underline{24.67}&\textbf{79.87} &24.14                  &67.13                  &51.51                  &\textbf{79.38} &29.59                  &64.16  & 4.94\\
\midrule
Ours                      &all     &\textbf{11.85} &\underline{78.60}&\textbf{43.99} &\underline{69.54}&\textbf{10.61} &77.12                  &\textbf{66.60} &\underline{67.32}  & 9.59\\
\bottomrule
\end{tabular}
}
\label{tab:controllability}
\end{table*}

\subsection{Main Results}
\label{subsec:Quantitative}
\noindent\textbf{Comparison of Controllability.}
As shown in Tab.\ref{tab:controllability}, we report the controllability comparison results across different datasets, where the structural similarity index (SSIM) is employed to evaluate image consistency. Our PixelPonder surpasses other methods with improvements of 8.41$\%$ and 19.29$\%$ on two benchmark datasets, establishing a new state-of-the-art in image generation. This notable accomplishment demonstrates that PixelPonder is capable of processing intricate combinations of multiple spatial conditions, generating images that are highly consistent with the given spatial constraints.

\noindent\textbf{Comparison of Image Quality.}
Previous methods often encounter two primary issues when handling multiple visual control conditions: a) Coarsening of lines within the image, and b) Disappearance or distortion of the main subject. In contrast, PixelPonder, leveraging its unique architectural design, effectively avoids these issues by harmonizing multiple visual control conditions without compromising image quality. This is demonstrated by the quantitative results in Tab.\ref{tab:controllability}, including significant improvements in FID (12.82 on MutiGen-20M and 20.34 on Subject-200K) and leading MUSIQ scores, and qualitative comparison in Fig.\ref{fig:vis_compare} further supports the effectiveness of PixelPonder in handling complex visual conditions.

\begin{wraptable}{r}{0.73\textwidth}
  \centering
  \caption{\textbf{Results of combining different condition types.} A quantitative comparison with models controlled by single visual conditions and various combinations of visual conditions. ``Source" refers to the original reference image.}
  \resizebox{1\linewidth}{!}{
    \begin{tabular}{c|c|c|c|c}
\toprule
\multicolumn{1}{c|}{} & \multicolumn{1}{c|}{}  & \multicolumn{3}{c}{\textbf{MultiGen-20M}}  \\ \cline{3-5}
\multicolumn{1}{c|}{\multirow{-2}{*}{\textbf{Methods}}}& \multicolumn{1}{c|}{\multirow{-2}{*}{\textbf{Conditions}}} &  \multicolumn{1}{c|}{\textbf{FID ($\downarrow$)}} & \multicolumn{1}{c|}{\textbf{SSIM~($\uparrow$)}} & \multicolumn{1}{c}{\textbf{MUSIQ ($\uparrow$)}} \\ 
\midrule
Source & - & -  & -  &  69.30 \\
ControlNet++ & canny & 17.69 & 36.69  &  65.67  \\
ControlNet++ & hed & 13.93 & 42.12  &  71.22 \\
ControlNet++ & depth & 17.56 & 27.79  &  71.23  \\ 
Ours & pose & 30.00 & 27.91  &  58.16  \\
Ours & pose+depth & 19.76 & 31.63  &  62.63 \\
Ours & pose+depth+hed & 12.41 & 40.48 & 67.82 \\
Ours & all & 11.85 & 43.99  &  69.54\\
\bottomrule
\end{tabular}
    }
   \vspace{-0.5cm}
  \label{tab:condition_types}%
\end{wraptable}
\noindent\textbf{Comparison of CLIP Score.}
Tab.\ref{tab:controllability} compares CLIP Score metric across methods under maximal conditions. The results reveal a consistent trade-off: methods with fewer visual constraints achieve higher text adherence. Specifically, PixelPonder demonstrates superior performance under full conditional control, outperforming the compared methods. Although Cocktail attains higher scores under significantly fewer visual constraints, the strong text adherence of Cocktail significantly compromises the consistency of its visual performance. As illustrated in Fig.\ref{fig:vis_compare}, the images generated by Cocktail exhibit pronounced issues of object loss and mutations when compared to those produced by other methods. This highlights PixelPonder’s unique capability to balance precise visual control with competitive text adherence.

\begin{figure*}[t]
    \centering
    \includegraphics[width=0.85\linewidth]{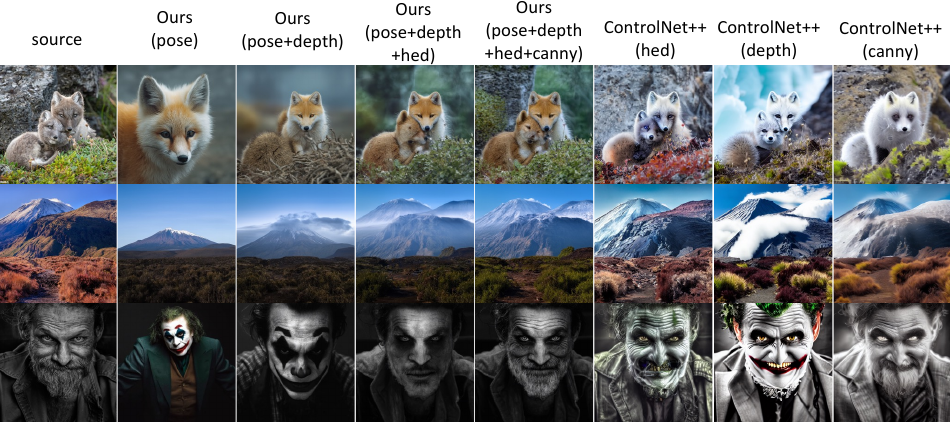}
    \caption{
         \textbf{Comparison of multi-condition image generation} by PixelPonder with different combinations against single-condition image generation by ControlNet++.
    }
    \label{fig:compare_little}
\end{figure*}
\noindent\textbf{Comparison of Inference Speed.} 
Tab.\ref{tab:controllability} presents the inference speed. 
Due to the fact that Cocktail involves only half the number of conditions compared to other methods, any direct comparison of time efficiency may not be entirely representative or meaningful. 
Patch Adaptation introduces a marginal increase in generation time; however, given the substantial improvement in image quality, the associated computational overhead is acceptable.

\noindent\textbf{Condition Types Comparison.}
To examine the effect of the number of condition types, Fig.\ref{fig:compare_little} shows the visual results under various condition combinations, with qualitative evaluation metrics provided in Tab.\ref{tab:condition_types}.  Examples in Fig.\ref{fig:compare_little} demonstrate that as the number of conditions increases, the layout and texture within images become increasingly refined and accurate, demonstrating that the control effects of different visual conditions are not identical. Furthermore, as can be seen from Tab.\ref{tab:condition_types}, an increase in visual conditions is beneficial for the overall quality and controllability of the images, which aligns with the observed visual effects. As evidenced by the comparison with ControlNet++~\citep{li2025controlnet}, integrating multi-conditions improves control precision and enables finer-grained manipulation of the generated output.

\begin{figure*}[t]
    \centering
    \includegraphics[width=1\linewidth]{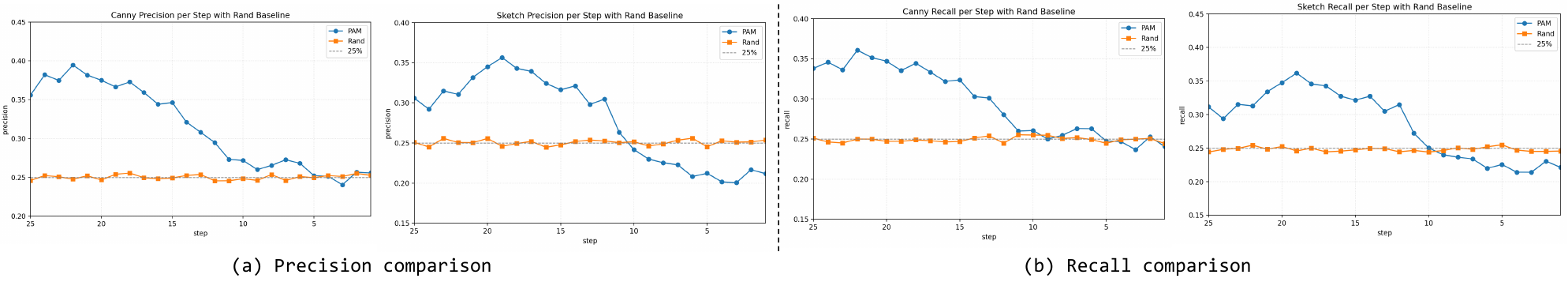}
    \caption{
         \textbf{Comparison of precision and recall} of PAM and random scheme on canny and sketch conditions. Taking Canny and Sketch as examples.
    }
    \label{fig:PR}
\end{figure*}
\begin{figure}[t]
    \centering
    \begin{minipage}{0.46\textwidth}
        \centering
        \small
        \captionsetup{type=table}
        \caption{\textbf{Ablation on zero controllable flow.} \texttimes \ represents the non-use of zero controllable flow, while the \checkmark \ represents the use of zero controllable flow.}
    \setlength{\tabcolsep}{2.4mm}
    \label{tab:Ablation_zero}
	\centering
	\resizebox{\linewidth}{!}{
	\begin{tabular}{c|c|c|c}
\toprule
\multicolumn{1}{c|}{} & \multicolumn{3}{c}{\textbf{MultiGen-20M}}  \\ \cline{2-4}
\multicolumn{1}{c|}{\multirow{-2}{*}{\textbf{Zero}}} &  \multicolumn{1}{c|}{\textbf{FID ($\downarrow$)}} & \multicolumn{1}{c|}{\textbf{SSIM~($\uparrow$)}} & \multicolumn{1}{c}{\textbf{MUSIQ ($\uparrow$)}}\\
\midrule
\texttimes   & 11.26 & 43.21  &  69.15  \\
\checkmark   & 11.85 & 43.99  &  69.54  \\
\bottomrule
\end{tabular}
	}
    \end{minipage}
    \hfill
    \begin{minipage}{0.48\textwidth}
        \centering
        \small
        \captionsetup{type=table}
        \caption{\textbf{Ablation on timestep information.} \texttimes \ represents the non-use of timestep information, while the \checkmark \ represents the use of timestep information.}
    \setlength{\tabcolsep}{1.5mm}
    \label{tab:Ablation_time}
	\centering
	\resizebox{\linewidth}{!}{
	\begin{tabular}{c|c|c|c}
\toprule
\multicolumn{1}{c|}{} & \multicolumn{3}{c}{\textbf{MultiGen-20M}}  \\ \cline{2-4}
\multicolumn{1}{c|}{\multirow{-2}{*}{\textbf{Timestep}}} &  \multicolumn{1}{c|}{\textbf{FID ($\downarrow$)}} & \multicolumn{1}{c|}{\textbf{SSIM~($\uparrow$)}} & \multicolumn{1}{c}{\textbf{MUSIQ ($\uparrow$)}}\\
\midrule
\texttimes   & 29.53 & 28.55  &  67.21  \\
\checkmark   & 11.85 & 43.99  &  69.54  \\
\bottomrule
\end{tabular}
	}
    \end{minipage}
\end{figure}


\subsection{Ablation Study}\label{subsec:ablation}
\begin{wrapfigure}{r}{0.49\linewidth}
    \centering
    \includegraphics[width=1\linewidth]{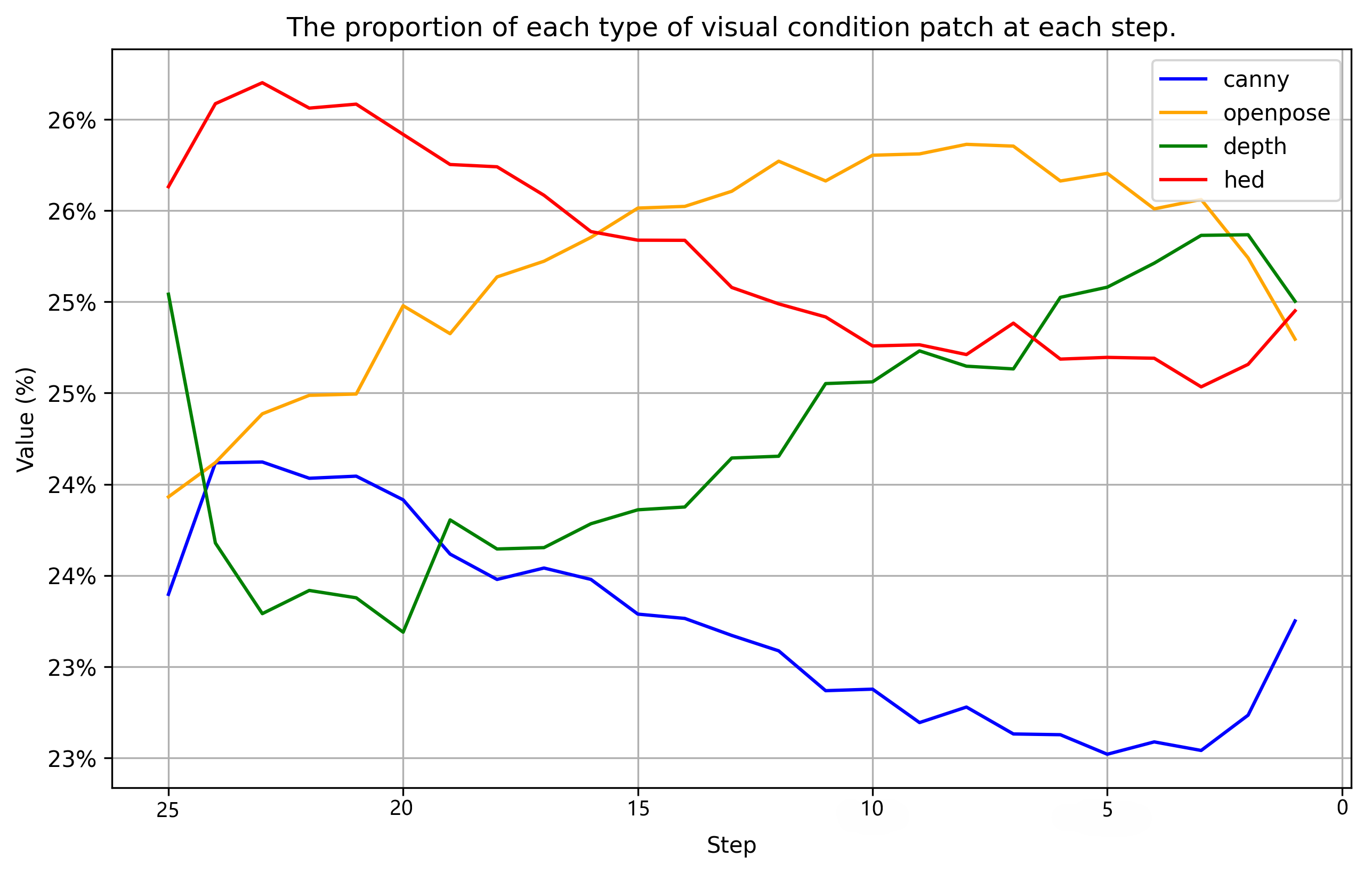}
    \caption{\textbf{Statistical distribution of each visual condition patch} during the denoising steps. }
    \label{fig:distribution}
\end{wrapfigure}

\noindent\textbf{Patch Selection.} In Fig. \ref{fig:distribution}, we illustrate the statistical distribution of each visual condition patch during the denoising steps. During the initial denoising stages from step 0 to 5, the denoising process comes from low-frequency regions to high-frequency regions. However, from step 5 to 25, the trend is completely opposite. This suggests that after establishing a rough layout in the initial stages, generating low-frequency or high-frequency regions does not necessarily require consistent information with the layout. On the contrary, it requires visual information from other frequencies to leave sufficient space to meet other visual requirements. More discussion is shown in the supplementary material.

\noindent\textbf{Spatial Correlation Analysis for Patch Selection Interpretability.} To quantitatively evaluate how well PAM’s selection aligns with local spatial energy, we constructed modality-specific importance reference maps. Taking Canny and Sketch as examples, (1) canny conditions: We computed gradient magnitude maps via Sobel operators (5,500 samples from MultiGen-20M). Images were partitioned into a patch grid matching PAM’s dimensions. After applying a global threshold, the top 25$\%$ of patches with the highest strong-edge pixel density were labeled as High-Density Edge Regions (Ground Truth), (2) sketch conditions: Similarly, importance was defined by line density (ratio of non-zero pixels). The top 25$\%$ densest patches were designated as the importance ground truth. As illustrated in Fig. \ref{fig:PR}, the random baseline maintains a stable precision of approximately 25$\%$ across all timesteps, which is expected given the uniform random selection across four condition types. In contrast, both Canny and Sketch exhibit a clear temporal evolution trend. This non-random, time-aware pattern demonstrates that PAM’s routing mechanism is not arbitrary. Its early-stage selection explicitly prioritizes spatially important regions, contributing to its empirical advantages over the random baseline.


\noindent \textbf{Zero Controllable Flow.} As shown in Tab.\ref{tab:Ablation_zero}, the use of zero controllable flow significantly enhances the aesthetic quality and color richness of the images. However, this comes with a slight loss of feature consistency in the images, while maintaining better visual coherence.


\noindent \textbf{Timestep Information Injection.} As shown in Tab.\ref{tab:Ablation_time}, controllable generation without timestep guidance exhibits a pronounced degradation in FID and SSIM, compared to generation with timestep guidance. This confirms that timestep information is essential for achieving effective structural controllability in generated images.
\section{Conclusion}
\label{sec:conclusion}
In this paper, we propose PixelPonder, a framework for compositional visual conditioning in diffusion models that addresses guidance redundancy in multi-control generation. Our key contributions include: a Patch-level Adaptive Condition Adaptation mechanism to resolve spatial conflicts via learnable attention gates, and a Time-aware Control Injection scheme to orchestrate condition influence across denoising phases. Experiments show significant improvements over state-of-the-art methods, enabling users to achieve precise, multi-aspect object creation through diverse visual conditions. With PixelPonder, users can depict different aspects of objects by utilizing various visual conditions, thereby realizing their creations.

\bibliography{neurips2026}

@inproceedings{li2025controlnet,
  title={ControlNet ++ : Improving Conditional Controls with Efficient Consistency Feedback},
  author={Li, Ming and Yang, Taojiannan and Kuang, Huafeng and Wu, Jie and Wang, Zhaoning and Xiao, Xuefeng and Chen, Chen},
  booktitle={European Conference on Computer Vision},
  pages={129--147},
  year={2025},
  organization={Springer}
}

@inproceedings{hu2023cocktail,
  title={Cocktail: Mixing multi-modality control for text-conditional image generation},
  author={Hu, Minghui and Zheng, Jianbin and Liu, Daqing and Zheng, Chuanxia and Wang, Chaoyue and Tao, Dacheng and Cham, Tat-Jen},
  booktitle={Thirty-seventh Conference on Neural Information Processing Systems},
  year={2023}
}

@article{zhao2024uni,
  title={Uni-controlnet: All-in-one control to text-to-image diffusion models},
  author={Zhao, Shihao and Chen, Dongdong and Chen, Yen-Chun and Bao, Jianmin and Hao, Shaozhe and Yuan, Lu and Wong, Kwan-Yee K},
  journal={Advances in Neural Information Processing Systems},
  volume={36},
  year={2024}
}

@article{sun2024anycontrol,
  title={AnyControl: Create Your Artwork with Versatile Control on Text-to-Image Generation},
  author={Sun, Yanan and Liu, Yanchen and Tang, Yinhao and Pei, Wenjie and Chen, Kai},
  journal={arXiv preprint arXiv:2406.18958},
  year={2024}
}

@article{qin2023unicontrol,
  title={Unicontrol: A unified diffusion model for controllable visual generation in the wild},
  author={Qin, Can and Zhang, Shu and Yu, Ning and Feng, Yihao and Yang, Xinyi and Zhou, Yingbo and Wang, Huan and Niebles, Juan Carlos and Xiong, Caiming and Savarese, Silvio and others},
  journal={arXiv preprint arXiv:2305.11147},
  year={2023}
}

@inproceedings{zhang2023adding,
  title={Adding conditional control to text-to-image diffusion models},
  author={Zhang, Lvmin and Rao, Anyi and Agrawala, Maneesh},
  booktitle={Proceedings of the IEEE/CVF International Conference on Computer Vision},
  pages={3836--3847},
  year={2023}
}

@article{dhariwal2021diffusion,
  title={Diffusion models beat gans on image synthesis},
  author={Dhariwal, Prafulla and Nichol, Alexander},
  journal={Advances in neural information processing systems},
  volume={34},
  pages={8780--8794},
  year={2021}
}

@inproceedings{rombach2022high,
  title={High-resolution image synthesis with latent diffusion models},
  author={Rombach, Robin and Blattmann, Andreas and Lorenz, Dominik and Esser, Patrick and Ommer, Bj{\"o}rn},
  booktitle={Proceedings of the IEEE/CVF conference on computer vision and pattern recognition},
  pages={10684--10695},
  year={2022}
}

@inproceedings{esser2024scaling,
  title={Scaling rectified flow transformers for high-resolution image synthesis},
  author={Esser, Patrick and Kulal, Sumith and Blattmann, Andreas and Entezari, Rahim and M{\"u}ller, Jonas and Saini, Harry and Levi, Yam and Lorenz, Dominik and Sauer, Axel and Boesel, Frederic and others},
  booktitle={Forty-first International Conference on Machine Learning},
  year={2024}
}

@inproceedings{peebles2023scalable,
  title={Scalable diffusion models with transformers},
  author={Peebles, William and Xie, Saining},
  booktitle={Proceedings of the IEEE/CVF International Conference on Computer Vision},
  pages={4195--4205},
  year={2023}
}

@inproceedings{mou2024t2i,
  title={T2i-adapter: Learning adapters to dig out more controllable ability for text-to-image diffusion models},
  author={Mou, Chong and Wang, Xintao and Xie, Liangbin and Wu, Yanze and Zhang, Jian and Qi, Zhongang and Shan, Ying},
  booktitle={Proceedings of the AAAI Conference on Artificial Intelligence},
  volume={38},
  number={5},
  pages={4296--4304},
  year={2024}
}

@article{cao2024controllable,
  title={Controllable generation with text-to-image diffusion models: A survey},
  author={Cao, Pu and Zhou, Feng and Song, Qing and Yang, Lu},
  journal={arXiv preprint arXiv:2403.04279},
  year={2024}
}

@article{ye2023ip,
  title={Ip-adapter: Text compatible image prompt adapter for text-to-image diffusion models},
  author={Ye, Hu and Zhang, Jun and Liu, Sibo and Han, Xiao and Yang, Wei},
  journal={arXiv preprint arXiv:2308.06721},
  year={2023}
}

@inproceedings{si2024freeu,
  title={Freeu: Free lunch in diffusion u-net},
  author={Si, Chenyang and Huang, Ziqi and Jiang, Yuming and Liu, Ziwei},
  booktitle={Proceedings of the IEEE/CVF Conference on Computer Vision and Pattern Recognition},
  pages={4733--4743},
  year={2024}
}

@article{he2024dynamiccontrol,
  title={DynamicControl: Adaptive Condition Selection for Improved Text-to-Image Generation},
  author={He, Qingdong and Peng, Jinlong and Xu, Pengcheng and Jiang, Boyuan and Hu, Xiaobin and Luo, Donghao and Liu, Yong and Wang, Yabiao and Wang, Chengjie and Li, Xiangtai and others},
  journal={arXiv preprint arXiv:2412.03255},
  year={2024}
}

@article{ho2020denoising,
  title={Denoising diffusion probabilistic models},
  author={Ho, Jonathan and Jain, Ajay and Abbeel, Pieter},
  journal={Advances in neural information processing systems},
  volume={33},
  pages={6840--6851},
  year={2020}
}

@article{song2020denoising,
  title={Denoising diffusion implicit models},
  author={Song, Jiaming and Meng, Chenlin and Ermon, Stefano},
  journal={arXiv preprint arXiv:2010.02502},
  year={2020}
}

@article{song2020score,
  title={Score-based generative modeling through stochastic differential equations},
  author={Song, Yang and Sohl-Dickstein, Jascha and Kingma, Diederik P and Kumar, Abhishek and Ermon, Stefano and Poole, Ben},
  journal={arXiv preprint arXiv:2011.13456},
  year={2020}
}

@inproceedings{ronneberger2015u,
  title={U-net: Convolutional networks for biomedical image segmentation},
  author={Ronneberger, Olaf and Fischer, Philipp and Brox, Thomas},
  booktitle={Medical image computing and computer-assisted intervention--MICCAI 2015: 18th international conference, Munich, Germany, October 5-9, 2015, proceedings, part III 18},
  pages={234--241},
  year={2015},
  organization={Springer}
}

@inproceedings{brooks2023instructpix2pix,
  title={Instructpix2pix: Learning to follow image editing instructions},
  author={Brooks, Tim and Holynski, Aleksander and Efros, Alexei A},
  booktitle={Proceedings of the IEEE/CVF Conference on Computer Vision and Pattern Recognition},
  pages={18392--18402},
  year={2023}
}

@article{hertz2022prompt,
  title={Prompt-to-prompt image editing with cross attention control},
  author={Hertz, Amir and Mokady, Ron and Tenenbaum, Jay and Aberman, Kfir and Pritch, Yael and Cohen-Or, Daniel},
  journal={arXiv preprint arXiv:2208.01626},
  year={2022}
}

@article{ramesh2022hierarchical,
  title={Hierarchical text-conditional image generation with clip latents},
  author={Ramesh, Aditya and Dhariwal, Prafulla and Nichol, Alex and Chu, Casey and Chen, Mark},
  journal={arXiv preprint arXiv:2204.06125},
  volume={1},
  number={2},
  pages={3},
  year={2022}
}

@article{chen2024pixart,
  title={Pixart-$\{$$\backslash$delta$\}$: Fast and controllable image generation with latent consistency models},
  author={Chen, Junsong and Wu, Yue and Luo, Simian and Xie, Enze and Paul, Sayak and Luo, Ping and Zhao, Hang and Li, Zhenguo},
  journal={arXiv preprint arXiv:2401.05252},
  year={2024}
}

@article{black2024flux,
  title={Flux: Official inference repository for flux.1models Accessed:2024-11-12.},
  author={Black Forest Labs.},
  journal={},
  year={2024}
}

@inproceedings{li2023gligen,
  title={Gligen: Open-set grounded text-to-image generation},
  author={Li, Yuheng and Liu, Haotian and Wu, Qingyang and Mu, Fangzhou and Yang, Jianwei and Gao, Jianfeng and Li, Chunyuan and Lee, Yong Jae},
  booktitle={Proceedings of the IEEE/CVF Conference on Computer Vision and Pattern Recognition},
  pages={22511--22521},
  year={2023}
}

@inproceedings{yang2023reco,
  title={Reco: Region-controlled text-to-image generation},
  author={Yang, Zhengyuan and Wang, Jianfeng and Gan, Zhe and Li, Linjie and Lin, Kevin and Wu, Chenfei and Duan, Nan and Liu, Zicheng and Liu, Ce and Zeng, Michael and others},
  booktitle={Proceedings of the IEEE/CVF Conference on Computer Vision and Pattern Recognition},
  pages={14246--14255},
  year={2023}
}

@inproceedings{chen2024training,
  title={Training-free layout control with cross-attention guidance},
  author={Chen, Minghao and Laina, Iro and Vedaldi, Andrea},
  booktitle={Proceedings of the IEEE/CVF Winter Conference on Applications of Computer Vision},
  pages={5343--5353},
  year={2024}
}

@inproceedings{xie2023boxdiff,
  title={Boxdiff: Text-to-image synthesis with training-free box-constrained diffusion},
  author={Xie, Jinheng and Li, Yuexiang and Huang, Yawen and Liu, Haozhe and Zhang, Wentian and Zheng, Yefeng and Shou, Mike Zheng},
  booktitle={Proceedings of the IEEE/CVF International Conference on Computer Vision},
  pages={7452--7461},
  year={2023}
}

@inproceedings{ruan2023mm,
  title={Mm-diffusion: Learning multi-modal diffusion models for joint audio and video generation},
  author={Ruan, Ludan and Ma, Yiyang and Yang, Huan and He, Huiguo and Liu, Bei and Fu, Jianlong and Yuan, Nicholas Jing and Jin, Qin and Guo, Baining},
  booktitle={Proceedings of the IEEE/CVF Conference on Computer Vision and Pattern Recognition},
  pages={10219--10228},
  year={2023}
}

@inproceedings{yang2023effective,
  title={Effective whole-body pose estimation with two-stages distillation},
  author={Yang, Zhendong and Zeng, Ailing and Yuan, Chun and Li, Yu},
  booktitle={Proceedings of the IEEE/CVF International Conference on Computer Vision},
  pages={4210--4220},
  year={2023}
}

@article{ranftl2020towards,
  title={Towards robust monocular depth estimation: Mixing datasets for zero-shot cross-dataset transfer},
  author={Ranftl, Ren{\'e} and Lasinger, Katrin and Hafner, David and Schindler, Konrad and Koltun, Vladlen},
  journal={IEEE transactions on pattern analysis and machine intelligence},
  volume={44},
  number={3},
  pages={1623--1637},
  year={2020},
  publisher={IEEE}
}

@article{canny1986computational,
  title={A computational approach to edge detection},
  author={Canny, John},
  journal={IEEE Transactions on pattern analysis and machine intelligence},
  number={6},
  pages={679--698},
  year={1986},
  publisher={Ieee}
}

@article{tan2024ominicontrol,
  title={OminiControl: Minimal and Universal Control for Diffusion Transformer},
  author={Tan, Zhenxiong and Liu, Songhua and Yang, Xingyi and Xue, Qiaochu and Wang, Xinchao},
  journal={arXiv preprint arXiv:2411.15098},
  year={2024}
}

@article{heusel2017gans,
  title={Gans trained by a two time-scale update rule converge to a local nash equilibrium},
  author={Heusel, Martin and Ramsauer, Hubert and Unterthiner, Thomas and Nessler, Bernhard and Hochreiter, Sepp},
  journal={Advances in neural information processing systems},
  volume={30},
  year={2017}
}

@article{wang2004image,
  title={Image quality assessment: from error visibility to structural similarity},
  author={Wang, Zhou and Bovik, Alan C and Sheikh, Hamid R and Simoncelli, Eero P},
  journal={IEEE transactions on image processing},
  volume={13},
  number={4},
  pages={600--612},
  year={2004},
  publisher={IEEE}
}

@inproceedings{radford2021learning,
  title={Learning transferable visual models from natural language supervision},
  author={Radford, Alec and Kim, Jong Wook and Hallacy, Chris and Ramesh, Aditya and Goh, Gabriel and Agarwal, Sandhini and Sastry, Girish and Askell, Amanda and Mishkin, Pamela and Clark, Jack and others},
  booktitle={International conference on machine learning},
  pages={8748--8763},
  year={2021},
  organization={PMLR}
}

@article{hessel2021clipscore,
  title={Clipscore: A reference-free evaluation metric for image captioning},
  author={Hessel, Jack and Holtzman, Ari and Forbes, Maxwell and Bras, Ronan Le and Choi, Yejin},
  journal={arXiv preprint arXiv:2104.08718},
  year={2021}
}

@inproceedings{ke2021musiq,
  title={Musiq: Multi-scale image quality transformer},
  author={Ke, Junjie and Wang, Qifei and Wang, Yilin and Milanfar, Peyman and Yang, Feng},
  booktitle={Proceedings of the IEEE/CVF international conference on computer vision},
  pages={5148--5157},
  year={2021}
}

@article{vaswani2017attention,
  title={Attention is all you need},
  author={Vaswani, A},
  journal={Advances in Neural Information Processing Systems},
  year={2017}
}

@inproceedings{xie2015holistically,
  title={Holistically-nested edge detection},
  author={Xie, Saining and Tu, Zhuowen},
  booktitle={Proceedings of the IEEE international conference on computer vision},
  pages={1395--1403},
  year={2015}
}

@article{podell2023sdxl,
  title={Sdxl: Improving latent diffusion models for high-resolution image synthesis},
  author={Podell, Dustin and English, Zion and Lacey, Kyle and Blattmann, Andreas and Dockhorn, Tim and M{\"u}ller, Jonas and Penna, Joe and Rombach, Robin},
  journal={arXiv preprint arXiv:2307.01952},
  year={2023}
}

@inproceedings{ramesh2021zero,
  title={Zero-shot text-to-image generation},
  author={Ramesh, Aditya and Pavlov, Mikhail and Goh, Gabriel and Gray, Scott and Voss, Chelsea and Radford, Alec and Chen, Mark and Sutskever, Ilya},
  booktitle={International conference on machine learning},
  pages={8821--8831},
  year={2021},
  organization={Pmlr}
}

@article{huang2023composer,
  title={Composer: Creative and controllable image synthesis with composable conditions},
  author={Huang, Lianghua and Chen, Di and Liu, Yu and Shen, Yujun and Zhao, Deli and Zhou, Jingren},
  journal={arXiv preprint arXiv:2302.09778},
  year={2023}
}

@article{cheng2023layoutdiffuse,
  title={Layoutdiffuse: Adapting foundational diffusion models for layout-to-image generation},
  author={Cheng, Jiaxin and Liang, Xiao and Shi, Xingjian and He, Tong and Xiao, Tianjun and Li, Mu},
  journal={arXiv preprint arXiv:2302.08908},
  year={2023}
}

@article{nichol2021glide,
  title={Glide: Towards photorealistic image generation and editing with text-guided diffusion models},
  author={Nichol, Alex and Dhariwal, Prafulla and Ramesh, Aditya and Shyam, Pranav and Mishkin, Pamela and McGrew, Bob and Sutskever, Ilya and Chen, Mark},
  journal={arXiv preprint arXiv:2112.10741},
  year={2021}
}

@article{zhang2023controllable,
  title={Controllable text-to-image generation with gpt-4},
  author={Zhang, Tianjun and Zhang, Yi and Vineet, Vibhav and Joshi, Neel and Wang, Xin},
  journal={arXiv preprint arXiv:2305.18583},
  year={2023}
}

@inproceedings{wang2024instancediffusion,
  title={Instancediffusion: Instance-level control for image generation},
  author={Wang, Xudong and Darrell, Trevor and Rambhatla, Sai Saketh and Girdhar, Rohit and Misra, Ishan},
  booktitle={Proceedings of the IEEE/CVF Conference on Computer Vision and Pattern Recognition},
  pages={6232--6242},
  year={2024}
}

@article{chen2023pixart,
  title={Pixart-alpha: Fast training of diffusion transformer for photorealistic text-to-image synthesis},
  author={Chen, Junsong and Yu, Jincheng and Ge, Chongjian and Yao, Lewei and Xie, Enze and Wu, Yue and Wang, Zhongdao and Kwok, James and Luo, Ping and Lu, Huchuan and others},
  journal={arXiv preprint arXiv:2310.00426},
  year={2023}
}

@article{bar2023multidiffusion,
  title={Multidiffusion: Fusing diffusion paths for controlled image generation},
  author={Bar-Tal, Omer and Yariv, Lior and Lipman, Yaron and Dekel, Tali},
  year={2023}
}

@article{jang2016categorical,
  title={Categorical reparameterization with gumbel-softmax},
  author={Jang, Eric and Gu, Shixiang and Poole, Ben},
  journal={arXiv preprint arXiv:1611.01144},
  year={2016}
}

@inproceedings{sohl2015deep,
  title={Deep unsupervised learning using nonequilibrium thermodynamics},
  author={Sohl-Dickstein, Jascha and Weiss, Eric and Maheswaranathan, Niru and Ganguli, Surya},
  booktitle={International conference on machine learning},
  pages={2256--2265},
  year={2015},
  organization={pmlr}
}
\bibliographystyle{abbrvnat}

\newpage
\renewcommand\thefigure{A\arabic{figure}}
\renewcommand\thetable{A\arabic{table}}  
\renewcommand\theequation{A\arabic{equation}}
\setcounter{equation}{0}
\setcounter{table}{0}
\setcounter{figure}{0}
\appendix
\onecolumn
\section*{Appendix}
\renewcommand\thesubsection{\Alph{subsection}}
\label{appendix}


\subsection{Reproducibility \& Impact Statement}
We have already elaborated on all the models or algorithms proposed, experimental configurations, and benchmarks used in the experiments in the main body or appendix of this paper. Furthermore, we declare that the entire code used in this work will be released after acceptance. Moreover, we plan to make the dataset and associated code publicly available for research. Nonetheless, we acknowledge the potential for misuse, particularly by those aiming to generate misinformation using our methodology. We will release our code under an open-source license with explicit stipulations to mitigate this risk.

\subsection{The Use of Large Language Models}
We use large language models solely for polishing our writing, and we have conducted a careful check, taking full responsibility for all content in this work.

\subsection{Implementation Details}\label{subsec:Implementation}
\textbf{Dataset.}
Our training dataset is randomly sampled from the MultiGen-20M~\citep{li2025controlnet} dataset, resulting in a training set of 2.5 million samples. Each training sample contains an input image \( x \), a text prompt, and a set of visual conditions \( \mathcal{C} = \{ C_1, C_2, \ldots, C_n \} \). This set \(\mathcal{C}\) of visual conditions includes various types of visual cues, such as canny maps, sketches, depth maps, and pose maps, obtained from several other methods~\citep{yang2023effective,ranftl2020towards,canny1986computational,xie2015holistically}. In the testing dataset, due to the small size of the MultiGen-20M test set, and to ensure fairness in comparison, we combine the test and validation sets of MultiGen-20M to form a test set of up to 5,500 text-image pairs. Additionally, to increase the confidence in our conclusions, we randomly sample 5,000 text-image pairs from the Subject-200K\citep{tan2024ominicontrol} dataset as a second test set. This dataset provides more complex text compared to MultiGen-20M, and the Subject-200Ks in each image are clearly defined.

\noindent \textbf{Training Setting.}
We conduct the training on 8 GPUs with 96GB of VRAM, using a batch size of 4 for every GPU, for a total of 200,000 training steps. The AdamW optimizer is employed with a learning rate of $2 \times 10^{-5}$.
The default setting for patch size is 2, and the default number of patches selected per \(U_i\) is set to \( \frac{W}{\text{patch size}} \) (where \( W \) is the width of the image features). The default depth of the ISB module is 1, and the default number of Zero DSBs and Zero SSBs is set to 2 and 4, respectively.
During training, we randomly select 3 out of 4 types of conditions (canny maps, depth maps, sketch maps, and pose maps) as visual control inputs. Regarding the pose condition, since not all images contain humans, we do not discard the pose condition to ensure its effective training.

\noindent \textbf{Inference Setting.}
During the inference process, the sampling step for PixelPonder is set to 25. For the comparative experiments with other methods, such as UnicontrolNet and Cocktail, the hyperparameters during inference is set to the official default configurations. In all visual displays except for the multiple generated images of the same object, the global random seed is set to 100. It is worth mentioning that assuming the visual conditions consist of \( H \times W \) tokens, the number of patches selected in each update process \( U_i \) is \( W \), and the total number of update processes is \( H \). 

\noindent\textbf{Evaluation Metrics.}
For multi-visual condition control tasks, we evaluate generation quality, controllability, and text-image consistency using a comprehensive set of mathematical and model-based metrics.\textbf{ Generation quality} is assessed through \textbf{FID}\citep{heusel2017gans} and \textbf{MUSIQ} \citep{ke2021musiq}. FID, a model-based metric, quantifies the similarity between feature distributions of real and generated images, providing a robust measure of image fidelity. MUSIQ, another perceptual model-based metric, evaluates image quality from a human perceptual perspective by capturing both local and global quality attributes. \textbf{Controllability} is measured using \textbf{SSIM} \citep{wang2004image}, a mathematically grounded metric that computes the structural similarity between reference and generated images, offering a precise assessment of visual fidelity. The evaluation metric for \textbf{inference speed} is the time cost per image, measured in seconds per image (s/per image). Finally, \textbf{text-image consistency} is evaluated via \textbf{CLIP Score} \citep{hessel2021clipscore,radford2021learning}, a model-based metric that measures the semantic alignment between input text and generated images.

\begin{table}[ht]
\centering
\caption{\textbf{Ablation on patch size.} We tested the impact of patch size in Patch Selection on controlling the generation.}
\label{tab:ablation_Patch_Size}
\setlength{\tabcolsep}{12pt}
\begin{tabular}{c|c|c|c|c}
\toprule
\multicolumn{1}{c|}{} & \multicolumn{4}{c}{\textbf{MultiGen-20M}}  \\ \cline{2-5}
\multicolumn{1}{c|}{\multirow{-2}{*}{\textbf{Patch Size}}} &  \multicolumn{1}{c|}{\textbf{FID ($\downarrow$)}} & \multicolumn{1}{c|}{\textbf{CLIP Score~($\uparrow$)}} & \multicolumn{1}{c|}{\textbf{SSIM~($\uparrow$)}} & \multicolumn{1}{c}{\textbf{MUSIQ ($\uparrow$)}}\\
\midrule
2   &11.26 &  78.62 & 43.21  & 69.15  \\
4   &12.80 &  78.24 & 40.73  & 68.43  \\
8   &13.46 &  78.07 & 39.75  & 68.15  \\ 
\bottomrule
\end{tabular}
\end{table}
\subsection{Patch Size} 
Tab.\ref{tab:ablation_Patch_Size} presents the performance of PixelPonder under different patch sizes. The experimental results indicate that finer-grained image condition combination can better resolve conflicts among various visual elements in the visual conditions.

\subsection{Scale Study}\label{subsec:scale}
To investigate whether PixelPonder demonstrates better image quality with larger parameter scales and training datasets, we evaluated the fidelity and consistency of the generated images using FID, SSIM and MUSIQ on different data scales and module depths.

\noindent \textbf{Data Scale.} 
We evaluate the performance of PixelPonder on different scales of training datasets, with the results shown in Tab.\ref{tab:data_scale}. Due to limitations in computational resources and time, we only explored up to 2.5 million data. As shown in the table, PixelPonder demonstrates significant improvements in FID and SSIM as the scale of the training dataset increases. However, MUSIQ, constrained by the original reference images, does not show further improvement after a substantial increase in the size of the data set. Compared to other methods, such as ControlNet++ and UniControl, our method achieves results comparable to those obtained by tens of millions of training samples, while being trained on only millions of samples. This indicates that there is still room for improvement in the precise control capabilities of PixelPonder.

\begin{table}[ht]
\centering
\caption{%
    \textbf{Ablation on data scale.} FID, SSIM and MUSIQ are reported on MultiGen-20M datasets. \( k \) represents thousand, and \( m \) represents million.
}
\setlength{\tabcolsep}{12pt}
\begin{tabular}{c|c|c|c}
\toprule
\multicolumn{1}{c|}{} & \multicolumn{3}{c}{\textbf{MultiGen-20M}}  \\ \cline{2-4}
\multicolumn{1}{c|}{\multirow{-2}{*}{\textbf{Data scale}}} &  \multicolumn{1}{c|}{\textbf{FID ($\downarrow$)}} & \multicolumn{1}{c|}{\textbf{SSIM~($\uparrow$)}} & \multicolumn{1}{c}{\textbf{MUSIQ ($\uparrow$)}}\\
\midrule
105k   & 12.20 & 40.79  &  68.93  \\
686k   & 11.43 & 41.13  &  69.16  \\
2.5m   & 11.26 & 43.21  &  69.15  \\ 
\bottomrule
\end{tabular}
\label{tab:data_scale}
\end{table}

\subsection{Patch Selection Analysis}\label{subsec:Number}
To further explore the specific internal processes of patch selection, we conduct some investigations, including the selection tendencies of patch selection for various visual conditions, as well as the characteristics of the denoising process based on patch selection.

Previous methods exhibit a progression of features from coarse to fine during the denoising process. Specifically, generative models first create the rough outlines and layouts of various objects, followed by the addition of details, which means generating from low-frequency regions to high-frequency regions. Generally, we believe that generating low-frequency visual elements requires low-frequency information, while high-frequency visual elements require high-frequency information. 

\begin{table}[ht]
\centering
\caption{%
    \textbf{Ablation on fourier.} \texttimes \ represents the non-use of fourier, while the \checkmark \ represents the use of fourier. 
}
\setlength{\tabcolsep}{12pt}
\begin{tabular}{c|c|c|c}
\toprule
\multicolumn{1}{c|}{} & \multicolumn{3}{c}{\textbf{MultiGen-20M}}  \\ \cline{2-4}
\multicolumn{1}{c|}{\multirow{-2}{*}{\textbf{Fourier}}} &  \multicolumn{1}{c|}{\textbf{FID ($\downarrow$)}} & \multicolumn{1}{c|}{\textbf{SSIM~($\uparrow$)}} & \multicolumn{1}{c}{\textbf{MUSIQ ($\uparrow$)}}\\
\midrule
\texttimes   & 12.79 & 44.97  &  68.38  \\
\checkmark   & 13.21 & 44.49  &  68.09  \\
\bottomrule
\end{tabular}
\label{tab:fourier}
\end{table}

To further validate our idea, we design a Fourier-based low-frequency and high-frequency signal correction, aiming at adjusting the intensity of low-frequency and high-frequency signals to be consistent with previous conclusions. Its mathematical form is as follows:

\begin{equation}
    f_{low} = \alpha M_{low} f t
\end{equation}

\begin{equation}
    f_{high} = \frac{\tilde{M}_{low} f}{min(t, \frac{2}{3})}
\end{equation}
where \( t \) is the current time step, \(M_{low}\) is a low-frequency mask, \(\tilde{M}_{low}\) is a high-frequency mask, \(\alpha\) is the correction hyperparameter, \( f \) is the spectrogram, and \( f_{\text{low}} \) and \( f_{\text{high}} \) represent the corrected low-frequency and high-frequency signals, respectively. \( f_{\text{low}} \) decreases over time steps, while \( f_{\text{high}} \) increases. We incorporate this Fourier correction into the control network, and the results are shown in Tab.\ref{tab:fourier}. It demonstrates that the image quality generated by the Fourier correction constructed under the previous conclusions has declined in various aspects, which indirectly supports the validity of our conclusions.

In the examples shown in Fig.4 
of the main paper, due to the fine-grained division of patches, the pose maps are mostly blank noise at many patch locations, resulting in a lack of control effect. However, contrary to our expectations, in the invalid areas corresponding to the pose maps, patch selection still chooses the blank noise from the pose maps to perturb the generation, which leads to better generation results. This is supported by the comparisons in Tab.\ref{tab:condition_types} 
of the main paper, where the generation results under all conditional control scenarios significantly outperform other combinations of conditions, particularly in consistency evaluation metrics such as FID and SSIM. This indicates that appropriately adding noise during the denoising process does not negatively impact image generation; rather, it frees up more space for other visual condition controls to exert their influence, which is consistent with our conclusions regarding the selection tendencies of various visual conditions in patch selection.

\subsection{Generation Stability Comparison}\label{subsec:Stability}
\begin{figure}[t]
    \centering
    \includegraphics[width=1\linewidth]{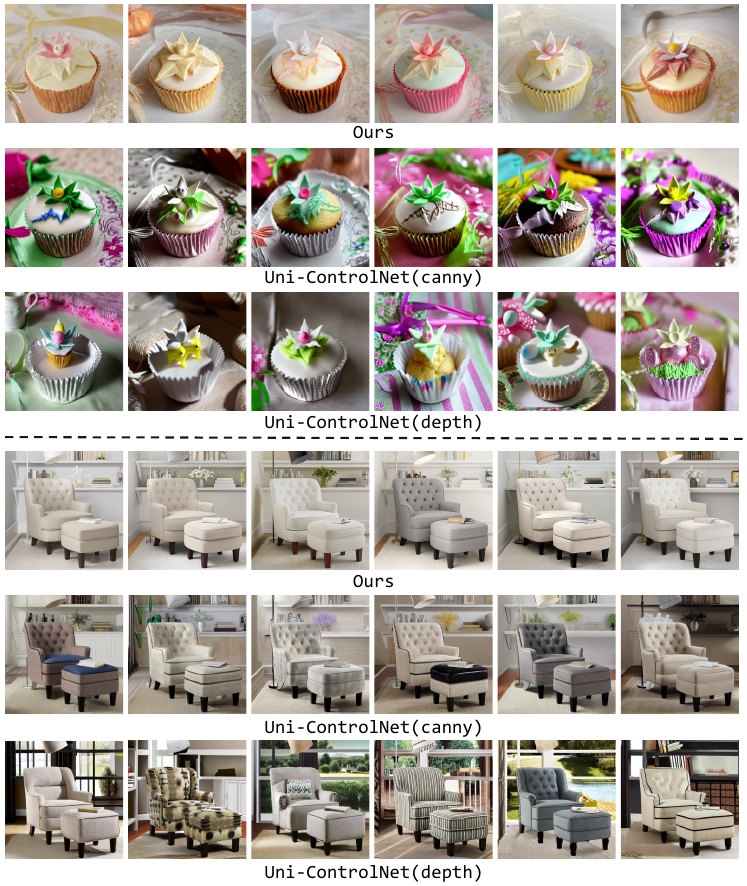}
    \caption{
         \textbf{Comparison of multiple random generations under single-condition and multi-condition.}  \textbf{\textit{Ours}} employs all available conditions, whereas \textbf{canny} and \textbf{depth} denote the use of only the canny or depth conditions, respectively.
    }
    \label{fig:Stability}
\end{figure}

\begin{figure}[h]
\centering
\includegraphics[width=0.95\linewidth]{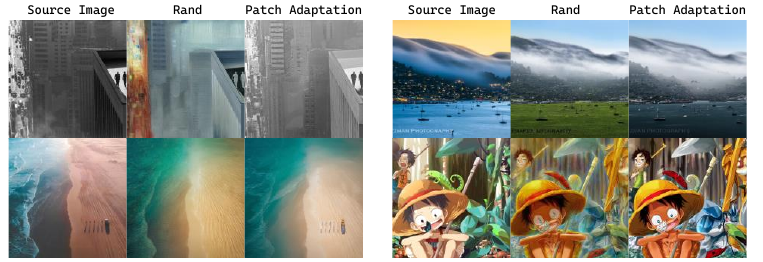}
\caption{\textbf{Generation quality comparison} between Random Selection and Patch Adaptation. Random selection can lead to a loss of information, resulting in reduced capability to generate details and clarity.}
\label{fig:comp_rand}
\end{figure}

As shown in Fig.\ref{fig:Stability}, the previous single-condition control framework encounters significant challenges in preserving both stability and consistency regarding the detailed features and morphological characteristics of target objects across multiple generated images. Particularly in the cake and furniture scenarios, Uni-ControlNet manifests several limitations, including texture degradation, contour deformation, and background inconsistency. In stark contrast, PixelPonder demonstrates superior performance in maintaining both consistency and stability across these dimensions. A notable example can be observed in the cake scenario, where PixelPonder effectively maintains precise control over the cream's morphological  features while simultaneously ensuring consistent dimensional attributes of the cake across all generated images.

As shown in Fig.\ref{fig:comp_rand}, randomly selected visual results exhibit significant deviations in terms of image layout, fine details, and intra-image contextual understanding of the garment-person interaction. Although the statistical difference is only a few percentage points, the visual discrepancy is highly noticeable. This demonstrates the effectiveness of the patch adaptation strategy in the generation process.

\subsection{Image Consistency Exploration}\label{subsec:Consistency}
\begin{figure*}[ht]
\centering
\includegraphics[width=0.95\linewidth]{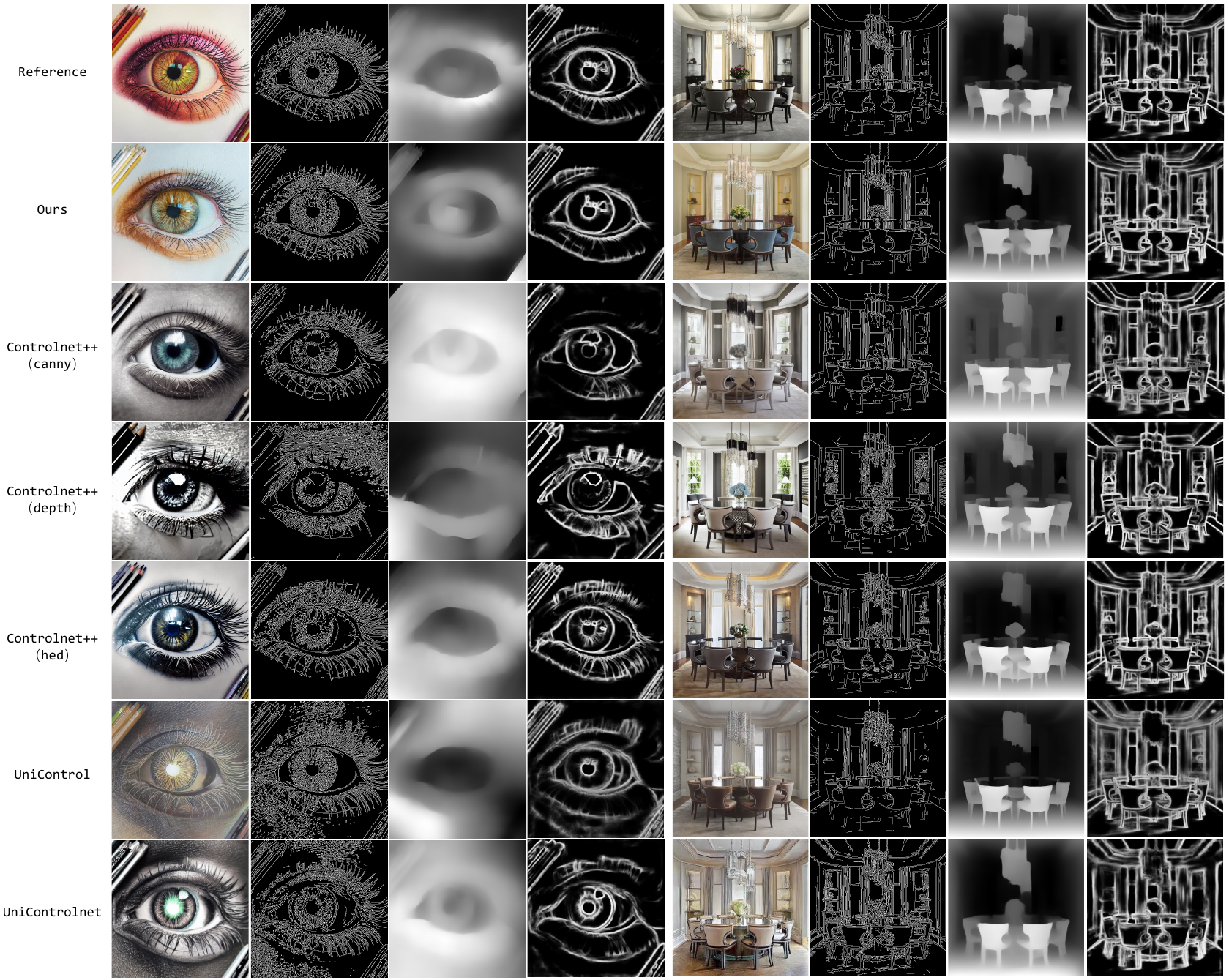}
\caption{\textbf{Image consistency comparison} between official or re-implemented methods and our proposed model.}
\label{fig:Consistency1}
\end{figure*}

\begin{figure*}[ht]
\centering
\includegraphics[width=0.95\linewidth]{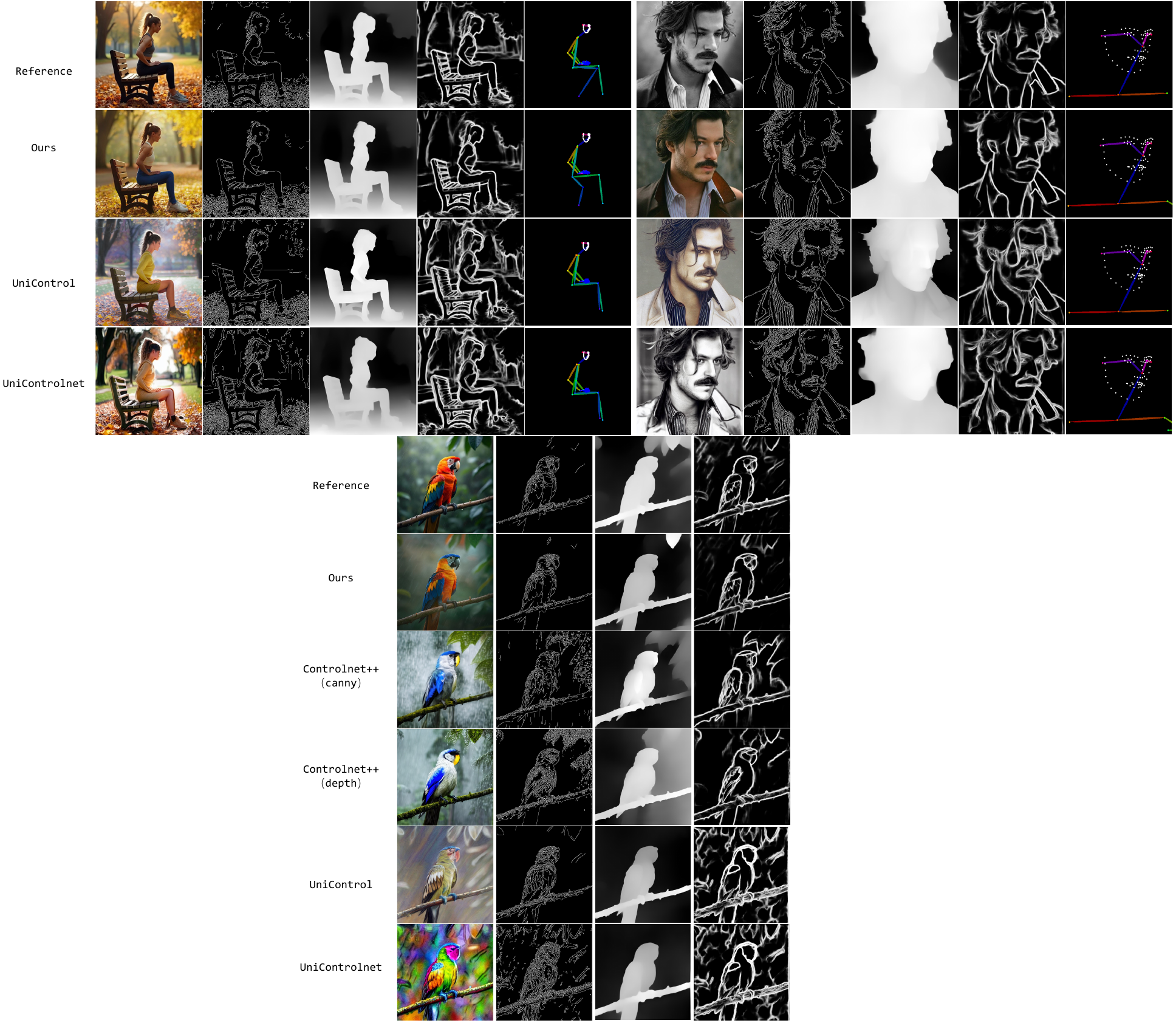}
\caption{\textbf{Image consistency comparison} between official or re-implemented methods and our proposed model.}
\label{fig:Consistency2}
\end{figure*}

To investigate whether the images generated by PixelPonder are consistent with the visual control conditions in terms of visual elements, we provide additional relevant visual demonstrations in Fig.\ref{fig:Consistency1} and Fig.\ref{fig:Consistency2}.

In each visual example, we present the reference along with its corresponding visual conditions. Additionally, the generated images are displayed under the visual conditions of the reference, as well as the extracted visual conditions from these generated images to investigate the consistency of each method with respect to various visual conditioning details in terms of visual fidelity and adherence to the specified conditions.

\begin{table}[h]
\caption{%
    \textbf{Comparison of image consistency and controllability on MultiGen-20M.} Note that \textbf{ControlNet++ is a single-condition model}, while other compared methods are \textbf{multi-condition models}. The diagonal cells (highlighted in gray) represent the optimal performance of ControlNet++ when evaluated under its corresponding visual condition. 
}
\resizebox{0.9\linewidth}{!}{
\begin{tabular}{c|c|c|c|c}
\toprule
\multicolumn{1}{c|}{} & \multicolumn{1}{c|}{}   & \multicolumn{3}{c}{\textbf{MultiGen-20M}}  \\ \cline{3-5}
\multicolumn{1}{c|}{\multirow{-2}{*}{Methods}}& 
\multicolumn{1}{c|}{\multirow{-2}{*}{Conditions}} &  
\multicolumn{1}{c|}{\makecell{Canny Edge (F1 Score$\uparrow$)}} & 
\multicolumn{1}{c|}{\makecell{Hed Edge (SSIM$\uparrow$)}} & 
\multicolumn{1}{c}{\makecell{Depth Map (RMSE$\downarrow$)}} \\ 
\midrule
ControlNet++ & canny & \cellcolor{gray!20}{31.98} & 49.61  &  14.58  \\
ControlNet++ & hed & 30.33 & \cellcolor{gray!20}{66.49}  &  12.17  \\ 
ControlNet++ & depth & 17.49 & 39.22  &  \cellcolor{gray!20}{11.60} \\
\midrule
T2I-Adapter & all & 11.35 & 28.93 & 24.97 \\
Uni-ControlNet & all & 21.97 & 41.96  &  17.14  \\
UniControl & all & \underline{26.8} & \underline{46.26} & \underline{13.26} \\
Ours & all & \textbf{28.35} & \textbf{60.51}  &  \textbf{11.93}\\
\bottomrule
\end{tabular}}
\label{tab:consistency_comparsion}
\end{table}

Tab.\ref{tab:consistency_comparsion} presents quantitative metrics evaluating the consistency between the condition maps of the generated images and those of the original reference images. Following the evaluation metrics of ControlNet++, we employ F1 Score for Canny edges, SSIM for HED edges, and RMSE for depth maps to assess consistency. The main comparison focuses on schemes with identical multi-visual conditions. Thus, ControlNet++ is excluded from direct comparison as it is a single-condition method.

Notably, \textbf{ControlNet++} represents the state-of-the-art single-condition control baseline within our knowledge. To investigate the performance of single-condition methods in terms of comprehensive control precision, we extracted multi-visual condition maps from images generated under single-condition constraints and calculated the corresponding metrics. As synthesized from Tab.~\ref{tab:consistency_comparsion} and Figs.~\ref{fig:Consistency1}-\ref{fig:Consistency2}, the \textit{specialization} of the single-condition scheme induces a significant \textbf{visual imbalance} phenomenon: while achieving exceptionally high consistency under the specific designated condition, its control precision for other unassigned conditions remains notably weak. This further highlights the critical need for coordinated multi-visual control tasks to achieve balanced and precise generation.

Moreover, multi-condition combination approaches (e.g., T2I-Adapter, Uni-ControlNet, UniControl) exhibit significantly inferior visual consistency compared to single-condition control methods overall. This indicates that existing multi-condition visual control methods struggle to harmonize the representations of diverse visual conditions, leading to degraded controllability as condition complexity increases.

PixelPonder, benefiting from its unique visual condition integration scheme, effectively leverages the complementary representational strengths of diverse visual conditions, achieving superior adherence across multiple scales (Canny, HED, and depth) compared to existing single- and multi-condition control approaches.

\subsection{More Qualitative Comparison}\label{subsec:More Visualization}
\begin{figure*}[ht]
\centering
\includegraphics[width=0.95\linewidth]{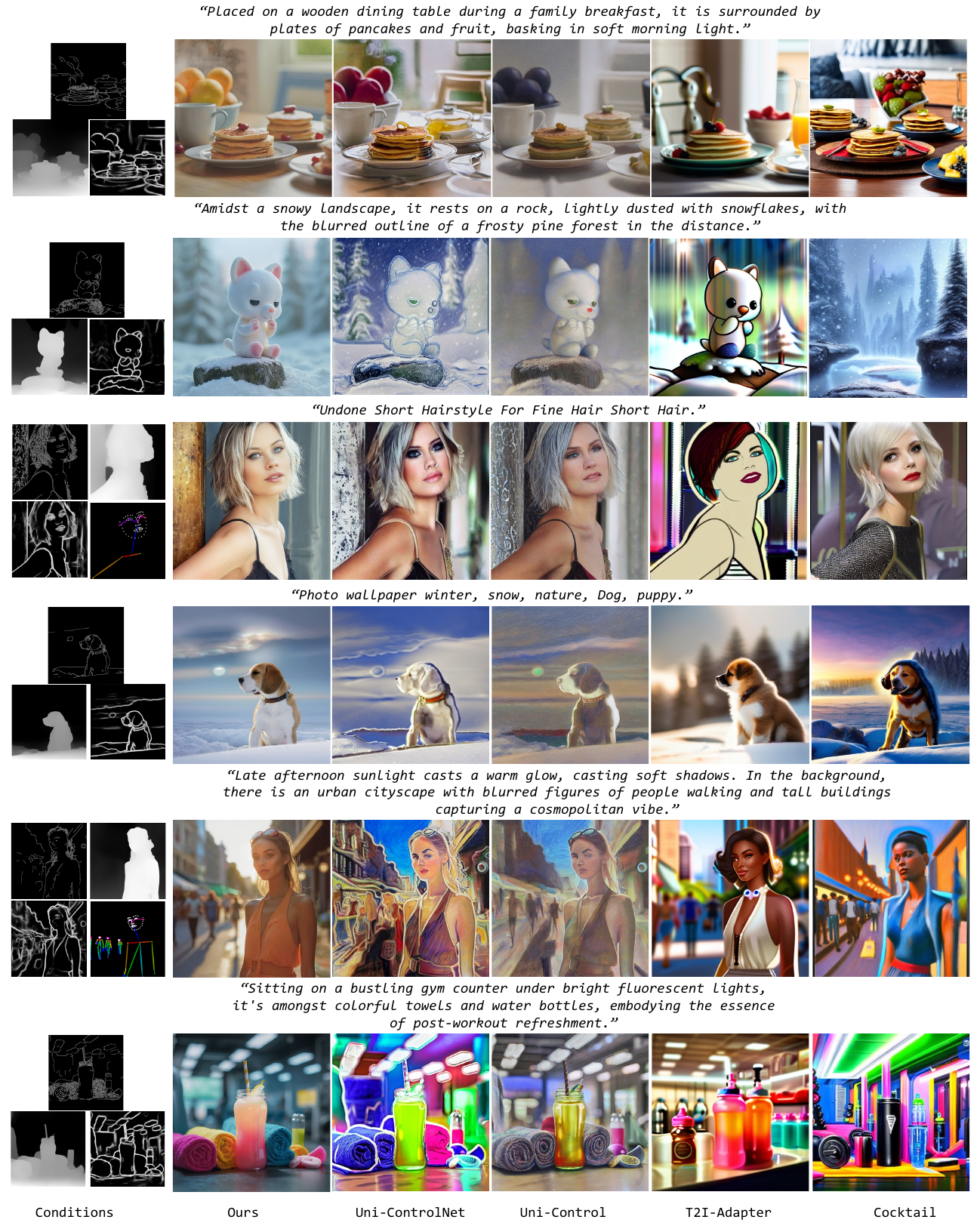}
\caption{\textbf{Image consistency comparison} between official methods and our proposed model.}
\label{fig:vis_supply_1}
\end{figure*}

\begin{figure*}[ht]
\centering
\includegraphics[width=0.95\linewidth]{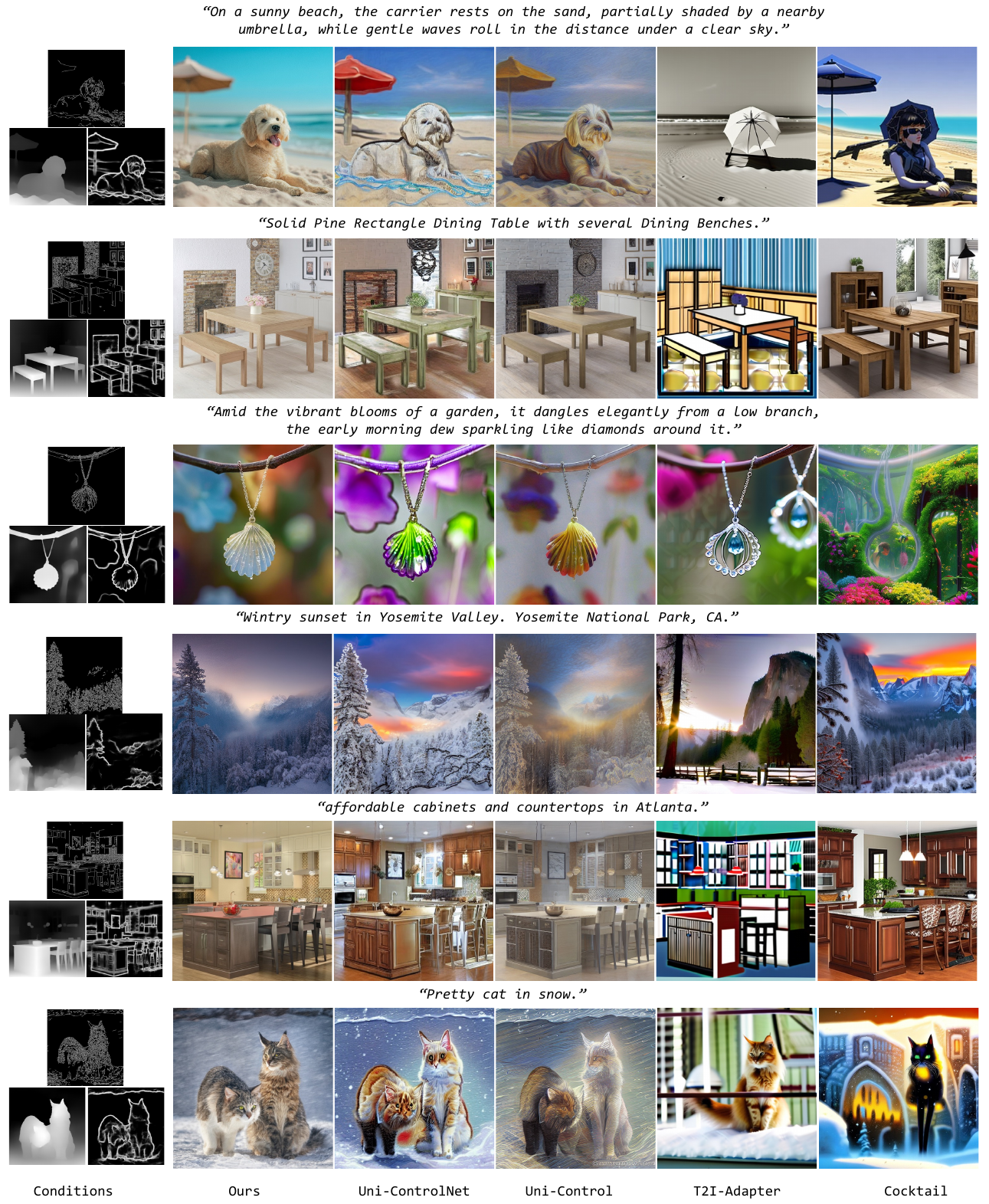}
\caption{\textbf{Image consistency comparison} between official methods and our proposed model.}
\label{fig:vis_supply_2}
\end{figure*}

In Fig.\ref{fig:vis_supply_1} and Fig.\ref{fig:vis_supply_2}, we present more visual results across various scenes. Previous methods do not separate multiple classes of visual conditions in visual control; instead, they inject multiple visual conditions simultaneously at a single time step, redundantly introducing visual elements of cannys. Our proposed PixelPonder, through patch-level separation of visual conditions and temporal domain separation, avoids the generation of overlapping conflicting visual signals and achieves the injection of complementary information among various visual elements in the temporal domain, resulting in superior generation performance under multi-visual condition control. 

\end{document}